\documentclass[10pt,journal,compsoc]{IEEEtran}
\usepackage{amsmath, amssymb}
\usepackage[dvipsnames,table]{xcolor}
\usepackage{floatflt}
\usepackage{multirow}
\usepackage{makecell}
\usepackage{booktabs}
\usepackage{url}
\usepackage{amsmath}
\usepackage{algorithm}
\usepackage{algpseudocode}
\usepackage{xcolor}
\ifCLASSOPTIONcompsoc
  \usepackage[nocompress]{cite}
\else
  \usepackage{cite}
\fi

\ifCLASSINFOpdf
   \usepackage[pdftex]{graphicx}
\else
\fi
\begin{document}
%


\title{Towards Efficient and Robust Linguistic Emotion Diagnosis for Mental Health via Multi-Agent Instruction Refinement}

%
%

\author{Jian~Zhang,
        Zhangqi~Wang,
        Zhiyuan~Wang,
        Weiping~Fu,
        Yu~He, \\
        Haiping~Zhu$^*$,
        Qika~Lin$^*$,~\IEEEmembership{Member,~IEEE},
        and~Jun~Liu,~\IEEEmembership{Senior Member,~IEEE}\thanks{* Corresponding author: Haiping~Zhu and Qika Lin.}

\IEEEcompsocitemizethanks{

\IEEEcompsocthanksitem Jian Zhang, Zhangqi Wang, Zhiyuan Wang, Weiping Fu, Yu He, Haiping Zhu and Jun Liu are with the School of Computer Science and Technology, Xi'an Jiaotong University, Shaanxi, China, 710049. E-mail: \{zhangjian062422, Asteria\_wzq, wang\_zy, fuweiping, 
heyucs\}@stu.xjtu.edu.cn, \{zhuhaiping, liukeen\}@xjtu.edu.cn. \protect\\

\IEEEcompsocthanksitem Qika Lin is with Saw Swee Hock School of Public Health, National University of Singapore, Singapore 119077. E-mail: qikalin@foxmail.com.}
}

%
%

\markboth{IEEE Transactions on Affective Computing 14(3), 2023
}%
{He \MakeLowercase{\textit{et al.}}: XX}
%



\IEEEtitleabstractindextext{%
\begin{abstract}
Linguistic expressions of emotions, including depression, anxiety, and trauma-related states, are widespread in clinical notes, counseling dialogues, and online mental health communities.
Accurate recognition of these emotions is crucial for clinical triage, risk assessment, and timely intervention in mental health related applications.
Despite recent advances showing that large language models (LLMs) can generalize well to various emotion analysis tasks, their diagnostic reliability in high-stakes and context-intensive medical settings remains highly sensitive to prompt design. 
Moreover, existing approaches are challenged by two major issues: \textit{emotional comorbidity}, where multiple intertwined emotional states complicate prediction, and \textit{inefficient exploration} of clinically relevant cues.
To address these issues, we propose \textbf{APOLO} (\textit{\underline{A}utomated \underline{P}rompt \underline{O}ptimization for \underline{L}inguistic em\underline{O}tion diagnosis}), a framework that systematically explores a broader and finer-grained prompt space to enhance diagnostic efficiency and robustness. APOLO models instruction refinement as a \textit{Partially Observable Markov Decision Process (POMDP)} and introduces a multi-agent collaboration mechanism comprising the \textit{Planner–Teacher–Critic–Student–Target} roles. This closed-loop design enables continuous optimization of prompt generation, evaluation, and evolution. After the \textit{Planner} agent formulates a high-level optimization trajectory within the POMDP framework, 
the \textit{Teacher–Critic–Student} agents collaboratively refine the prompts along this trajectory, 
iteratively enhancing the stability and effectiveness of the reasoning process.
Finally, the \textit{Target} agent determines whether to continue optimization or terminate the search 
based on performance evaluation. 
Experimental results demonstrate that APOLO improves diagnostic accuracy and robustness 
across domain-specific and stratified benchmarks, providing a generalizable and scalable paradigm 
for trustworthy LLM applications in mental healthcare.
\end{abstract}

}

\maketitle

\IEEEdisplaynontitleabstractindextext

%
\IEEEpeerreviewmaketitle

\IEEEraisesectionheading{\section{Introduction}\label{sec:introduction}}

Emotion diagnosis for mental health plays an essential role in understanding mental health conditions and tracking the progression of related disorders~\cite{kumar2024computational}. 
Emotions associated with illnesses, including depression, anxiety, and posttraumatic stress, frequently appear in clinical notes, counseling conversations, and online mental-health communities~\cite{ge2025survey,zhang-physreason}.
These texts capture patients’ cognitive and affective states and play a vital role in clinical triage, risk assessment, and intervention decisions~\cite{bucur2025survey}. 
Traditional emotion recognition methods, often based on shallow features or lexicons, perform reasonably well on generic sentiment tasks but struggle in medical contexts due to semantic ambiguity, implicit emotional expressions, and domain variability~\cite{zhang2023emotion,fu2026erreval,yan2025mur}.

\begin{figure}[t]
 \includegraphics[width=\columnwidth]{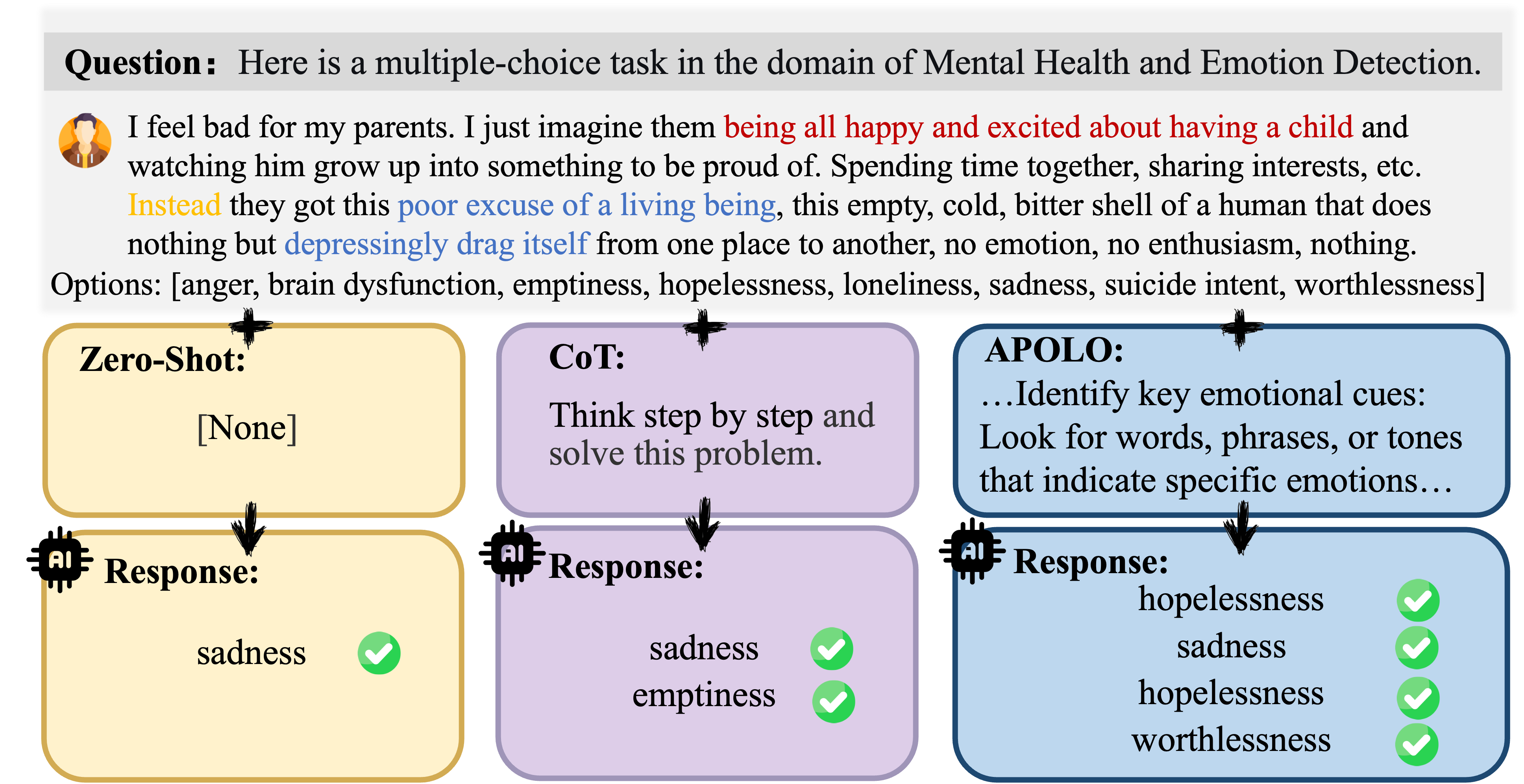}
 \caption{Examples of disease-related emotion diagnosis under three prompting strategies: zero-shot, CoT, and APOLO.}
 \label{fig_example}
\end{figure}

Recently, large language models (LLMs) have demonstrated remarkable capabilities in affective and psycholinguistic reasoning, achieving strong few-shot or zero-shot performance on mental health–related text analysis~\cite{sahoo2024systematic,xu2025large,lin2025has}. 
However, in high-stakes clinical scenarios, LLM outputs are highly sensitive to the phrasing and structure of prompts. 
Minor variations in instruction design or reasoning path can lead to drastically different diagnostic outcomes~\cite{li2025survey}. 
For instance, “I can’t sleep lately and don’t want to see anyone” implies both anxiety and depression, yet a poorly designed prompt may capture only one. 
Hand-crafted prompts, although guided by expert intuition, fail to systematically cover diverse semantic cues and implicit intentions, resulting in limited consistency and transferability~\cite{li2025survey,cui2025automatic}.

\begin{figure*}[t]
	\large
	\centering
	\includegraphics[scale=0.5]{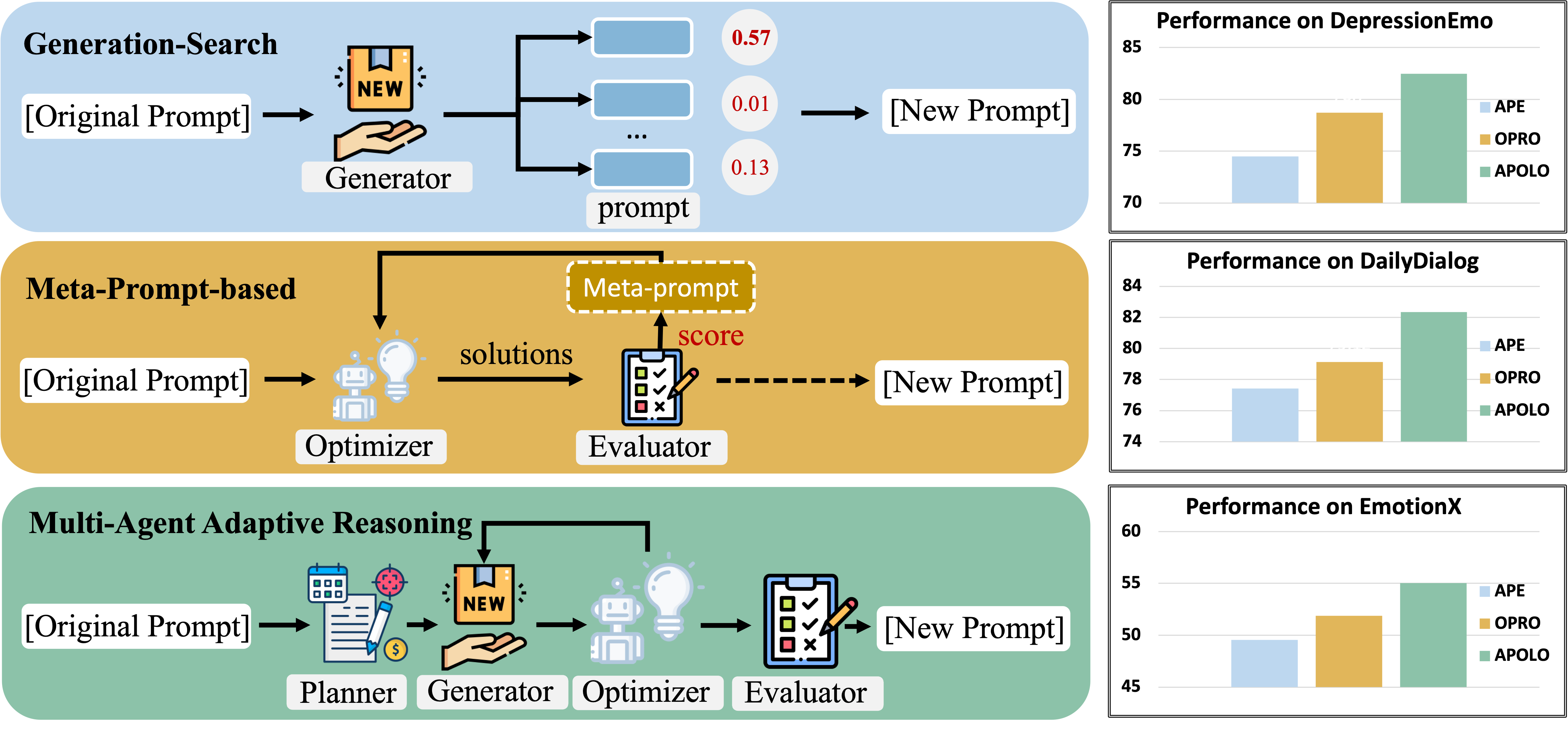}
	\caption{
Comparison of APO strategies. Top: generation–search methods generate and locally refine candidate prompts, leading to limited coverage. 
Middle: meta-prompt approaches rely on fixed optimization templates with low adaptability. 
Bottom: our APOLO framework introduces multi-agent adaptive reasoning, enabling dynamic and collaborative prompt exploration. 
Right: performance comparison across three datasets based on m-F1 score.}
	\label{compare}
\end{figure*}

Automated Prompt Optimization (APO) provides a principled way to overcome these limitations by systematically exploring a broader and finer-grained prompt space~\cite{ramnath2025systematic}. 
Through iterative search and evaluation, APO enables adaptive discovery of optimal task-specific prompts~\cite{li2025survey}. 
Nevertheless, generic APO methods remain inadequate for emotion diagnosis, where complex high-context semantics and safety constraints prevail~\cite{chang2024efficient}. 
As shown in Figure~\ref{fig_example}, we compare three prompting strategies for disease-related emotion recognition. 
The zero-shot prompt identifies only one emotion, while the Chain-of-Thought (CoT) prompt captures two, but both yield incomplete emotion diagnoses. 
This observation highlights that conventional prompting strategies fail to uncover the full spectrum of disease-related emotions, motivating the need for a more systematic and adaptive approach to prompt optimization~\cite{hegde2025emotions}. 

Such incompleteness arises from two fundamental issues: \textbf{emotional comorbidity} and \textbf{inefficient exploration}.  

\textbf{Emotional comorbidity.}
Disease-related emotions often appear in intertwined and co-occurring forms, such as anxiety with depression, trauma with guilt, or anhedonia with self-blame. 
A single utterance may convey multiple affective states that interact hierarchically or causally. 
Fixed-template prompts or single-label formulations fail to capture such dependencies~\cite{kumar2024computational}. 
For instance, the sentence “I feel guilty and afraid that I might break down” simultaneously expresses guilt and fear; without hierarchical reasoning, the model tends to detect only the dominant emotion. 
Existing prompt designs typically rely on flat labels or keyword cues, lacking mechanisms to model emotional co-occurrence and semantic overlap. 
As shown in Figure~\ref{fig_example}, both zero-shot and CoT prompts fail to capture the full set of coexisting emotions, whereas our framework dynamically adjusts prompt structure to mitigate label ambiguity and missing detection caused by comorbidity.

\textbf{Inefficient exploration.}
Most existing APO methods rely on two main strategies, namely \textit{generation–search} and \textit{meta-prompt}, as illustrated in Figure~\ref{compare}. 
Generation–search methods~\cite{zhou2023large,xureprompting,wang2023promptagent} generate candidate prompts and refine them through local search~\cite{ramnath2025systematic}, 
but they typically explore only a limited region of the prompt space and are prone to premature convergence~\cite{cui2025automatic}. 
Meta-prompt approaches~\cite{Yang0LLLZC24,ye2023prompt} design high-level instruction templates to guide optimization, 
but these templates are rigid and fail to adapt dynamically to diverse task contexts. 
Both paradigms lack global planning, uncertainty modeling, and interactive feedback, which are essential for effective search in high-dimensional prompt spaces. 
Such inefficiency not only constrains optimization performance but also increases instability, 
which is particularly critical in safety-sensitive medical applications.

To address the above issues, we propose \textbf{APOLO} (\textit{Automated Prompt Optimization for Linguistic Emotion Diagnosis}), 
which formulates prompt optimization as a \textit{Partially Observable Markov Decision Process (POMDP)} and introduces a multi-agent collaboration mechanism consisting of 
the \textbf{Planner–Teacher–Critic–Student–Target} roles.  
As illustrated in Figures~\ref{fig_example} and~\ref{compare}, APOLO dynamically explores the prompt space through adaptive reasoning and collaborative feedback, 
enabling the model to capture a more complete set of emotions with greater interpretability and stability.  
Specifically, the \textit{Planner} designs task-aware search trajectories under risk and cost constraints; 
the \textit{Teacher–Critic–Student} trio refines reasoning chains via Socratic-style dialogue, 
providing interpretable pseudo-gradient feedback; 
and the \textit{Target} agent evaluates performance and safety metrics to complete the optimization loop. 
Our main contributions are summarized as follows:


$\bullet$ \textbf{APOLO framework.} We introduce a POMDP-based multi-agent architecture that enables dynamic and interpretable instruction refinement for linguistic emotion diagnosis.  

$\bullet$ \textbf{Task-specific optimization.}  
We incorporate risk-aware and cost-constrained planning to alleviate emotion-diagnosis–specific challenges such as emotional comorbidity and uncertain inference.  

$\bullet$ \textbf{Comprehensive validation.}  
Extensive experiments demonstrate the effectiveness, robustness, and scalability of the modified model across diverse linguistic and clinical scenarios.

This paper is an extension of our previous conference work~\cite{zhang2026mars}. 
Compared with that publication, we have made the following improvements: 
(1) this study extends the MARS framework to the Linguistic Emotion Diagnosis task, providing a detailed analysis of its unique issues and corresponding solutions; 
(2) in Section~\ref{sec:method}, we introduce a risk-aware and cost-constrained planning module tailored to mitigate emotion-diagnosis–specific issues; 
and (3) more comprehensive experiments are conducted to verify the effectiveness and robustness of the modified model across various datasets.

\section{Related Work \label{sec:Related Work} }
This section aims to review two research areas closely related to this paper: Linguistic Emotion Diagnosis and APO. We review and comment on the core research in these two fields, respectively, to lay the theoretical and practical foundation for the methods proposed in this paper.

\subsection{Linguistic Emotion Diagnosis}

Linguistic emotion diagnosis is a core subfield of affective computing that aims to identify and interpret human emotions from textual signals. Distinct from broader sentiment analysis (which often emphasizes polarity), emotion diagnosis typically addresses finer-grained emotion categories or dimensions (e.g., anger, joy, sadness, fear) and the temporal dynamics of affective states. Early methods relied on lexicons and rule-based approaches, followed by classical supervised machine-learning models such as naive Bayes and SVMs\cite{singh2023detailed,hassan2023comprehensive,he2022jcbie}; these approaches achieved reasonable performance on certain benchmarks but commonly required extensive feature engineering and showed limited cross-domain transferability.


To better capture temporal dependencies in text, sequence models based on deep learning, such as RNNs, LSTMs, and their variants, are applied to emotion recognition\cite{zhu2024enhancing, bhat2021cnn, mishra2020speech,lan2025gem}, delivering advantages for sentence-level and discourse-level dynamics modeling. Some works extend CNN-LSTM architectures to multimodal settings (e.g., combining text with speech or facial cues), which enhance emotion recognition by integrating prosodic and visual signals; such studies should be distinguished from pure textual methods.

More recently, pre-trained language models (PLMs) built on the Transformer architecture (e.g., BERT) have substantially advanced text-based emotion detection by providing contextualized representations learned from large unlabeled corpora. Reviews of BERT-based methods\cite{acheampong2021transformer,bao2021bert} report significant gains in capturing implicit emotion and contextual relationships, while also noting limitations in few-shot scenarios, long-tail emotion classes, and cross-domain generalization.

Beyond static classification, recent research has focused on emotion dynamics, the patterns of how an individual’s emotions evolve over time, and explored their potential as linguistic biosocial markers for mental health. Empirical studies, such as those by Vishnubhotla and Mohammad\cite{vishnubhotla2023language}, demonstrate that statistics such as mean valence, variability, and recovery rate derived from textual sequences correlate significantly with conditions such as depression and PTSD , suggesting avenues for early, non-invasive mental-health screening via language signals.

Nonetheless, challenges remain: annotation subjectivity and inter-annotator agreement for fine-grained emotions, cross-lingual and cross-cultural generalization, class imbalance and rare emotions, and the high cost of reliable evaluation. Recent directions include leveraging PLM transfer and fine-tuning, weakly-/distantly-supervised and few-shot learning to reduce annotation needs, and integrating temporal dynamics modeling to produce more robust individual affective profiles. In this work we build on these trends and emphasize sample-efficient, semantically sensitive approaches (e.g., structured prompts combined with task-adaptive strategies) to improve fine-grained emotion discrimination in linguistically subtle contexts.

\subsection{APO}

APO seeks to enhance the downstream performance of LLMs by automatically refining prompts. Early APO work can be broadly categorized into two types: one focusing on discrete hard prompts (token-level) such as AutoPrompt\cite{shin2020autoprompt}, and another focusing on continuous soft prompts (vector-level) such as Prefix-Tuning\cite{li2021prefix}\cite{zhang2025gkg}. The former replaces tokens within predefined templates via gradient or enumeration methods to trigger latent model knowledge, while the latter prepends trainable vectors to input sequences, keeping the model body frozen for task adaptation. Although these approaches achieved notable results in some settings, they are often highly dependent on specific tasks or prompt templates, and tend to suffer from local-optima or limited transferability in the context of large-scale models, semantically rich tasks, or high evaluation cost.

With the proliferation of powerful LLMs, research emphasis has shifted toward semantic-level prompt optimization. The APE (Automatic Prompt Engineer) framework exemplifies this shift: it uses an LLM to generate candidate natural-language instructions, then filters and selects optimal prompts via scoring and search\cite{zhou2023large,zhang2026maxs}. Building on this, prompt optimization research has coalesced into two major strands: first, the generate-and-search paradigm, which generates multiple candidate prompts and then applies search, evolutionary or Monte Carlo mechanisms for ranking, filtering, or iterative refinement (e.g., PromptAgent\cite{wang2023promptagent}); second, the meta-prompting strategy, which designs “higher-order prompts that generate or improve other prompts”\cite{zhang2024meta}. Meta-prompts provide structured scaffolds or templates, enabling LLMs to fill in task-specific details and hence generate efficient prompts, thereby improving generality and design efficiency.

\begin{figure*}[t]
	\large
	\centering
	\includegraphics[scale=0.34]{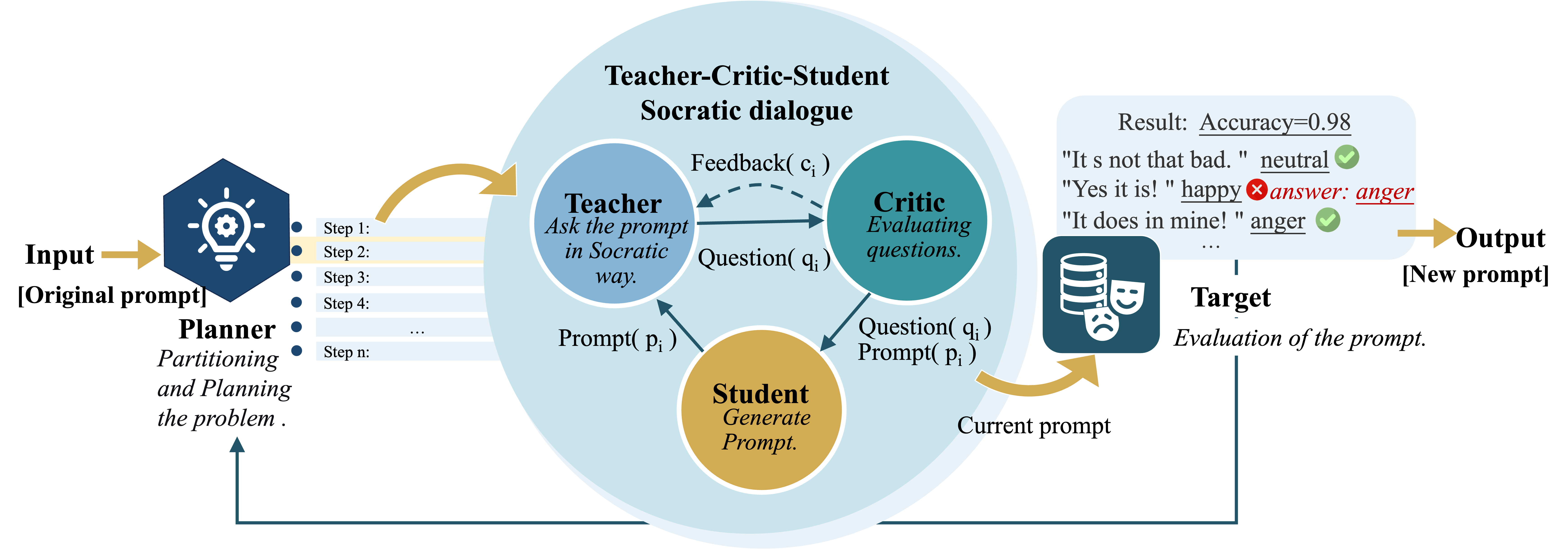}
	\caption{
  The overall architecture of the APOLO model, designed to support medical emotion diagnosis tasks.
  It consists of five LLM agents. The \textit{Planner} agent that autonomously generates task-specific optimization trajectories, and a \textit{Teacher}-\textit{Critic}-\textit{Student} Socratic dialogue mechanism that iteratively refines prompts, with the evaluation and iterative refinement process guided by feedback from the \textit{Target} agent.}
	\label{fig_model}
\end{figure*}

More recent work casts prompt optimization as a decision or control problem: for instance, RLPrompt formulates discrete prompt generation as a reinforcement-learning agent process, optimizing prompt policies with task-specific reward signals\cite{deng2022rlprompt}; moreover, multi-role/multi-agent frameworks (generator–evaluator–critic) \cite{zhang2025maps} explore iterative conversational loops across prompt generation, evaluation and rewriting stages to evolve high-quality prompts\cite{zhang2026mars}. These trends suggest that APO is transitioning from token-level search toward higher-level, structured, self-reflective optimization mechanisms. Given that linguistic emotion-diagnosis tasks involve subtle affective expressions, strong context dependence, and expensive annotation/assessment, this paper builds on that lineage to propose an APO method combining structured meta-prompts with task-adaptive optimization for this domain.

\section{Methodology}
\label{sec:method}

This section describes the overall APOLO framework, consisting of:
(1) the task formulation, 
(2) risk-aware and cost-constrained hierarchical trajectory planning,
(3) Socratic-style joint policy refinement,
(4) evaluation and convergence procedures, and
(5) theoretical insights together with the full optimization algorithm.

\subsection{Task Formulation}
\label{sec:taskdefinition}

In linguistic emotion diagnosis, the target model 
$\mathcal{\pi}_{\text{tar}}$ receives a text instance $x$ and must identify 
a set of co-occurring emotional states.  
Given an initial instruction $p_0$, APOLO aims to construct an optimized 
instruction $p^*$ that yields reliable multi-label predictions across 
emotion categories.  
Each sample $(x, Y)$ consists of a clinical or psychological text $x$ and 
a multi-label emotion set $Y \subseteq \mathcal{E}$ reflecting possible 
emotional comorbidity.

To quantify optimization performance, we define the objective:
\begin{equation}
p^* = 
\arg\max_{p\in\mathcal{P}}
\; \mathbb{E}_{(x,Y)\sim D_{\text{test}}}
\big[
f\!\left(\mathcal{\pi}_{\text{tar}}(x \mid p),\, Y\right)
\big],
\label{eq:objective}
\end{equation}
where $f(\cdot)$ denotes multi-label evaluation metrics such as 
Micro F1 or Jaccard similarity.  
This objective captures the goal of improving both completeness and 
consistency in emotion detection.

In line with the iterative nature of clinical assessment, APOLO models the 
multi-agent refinement process as a partially observable decision system 
$(\mathcal{S},\mathcal{A},\mathcal{T},\mathcal{O},\mathcal{R})$, 
where $\mathcal{S}$ reflects latent diagnostic cues and uncertainty derived 
from the patient's narrative, $\mathcal{A}$ corresponds to 
clinically motivated interventions generated by the Teacher–Critic agents 
(e.g., probing for comorbid signals or clarifying ambiguous affect), 
$\mathcal{T}$ updates the internal diagnostic hypothesis, 
$\mathcal{O}$ expresses this latent state as an actionable instruction for 
the target model, and 
$\mathcal{R}=f(\mathcal{\pi}_{\text{tar}}(x\mid\mathcal{O}(s)),Y)$ 
quantifies diagnostic utility.  
This formulation enables APOLO to mirror real-world clinical reasoning: 
prompts evolve through uncertainty-aware state transitions rather than 
direct text editing, allowing the system to account for subtle emotional 
co-occurrence and risk-related linguistic cues.


\subsection{Risk- and Cost-Aware Trajectory Planning}
\label{sec:plan}

The \textit{Planner} is responsible for generating a sequence of intermediate
sub-goals that guides the downstream multi-agent refinement.
Given task goal $g$, input $x$, and initial prompt $p_0$, it outputs a trajectory
\(
\mathbf{ST} = [st_1,\ldots,st_n] = \pi_{\text{plan}}(g, x, p_0),
\)
where each $st_i$ specifies a localized reasoning focus (e.g., identifying
candidate emotions, probing uncertainty, or clarifying clinical details).

A Conventional trajectory planner selects $\mathbf{ST}$ by maximizing its
likelihood under a latent semantic context $z$ drawn from $q(z\mid g,x)$:
\begin{equation}
\mathbf{ST}_{\text{base}}
= \arg\max_{\mathbf{ST}}
\; \mathbb{E}_{z\sim q(z\mid g,x)}
\big[
\log P(\mathbf{ST}\mid z,p_0)
\big].
\label{eq:planner_base}
\end{equation}
This objective encourages trajectories that are semantically plausible and
aligned with contextual cues, but it lacks explicit mechanisms for handling
clinical risk or computational efficiency.

In emotion diagnosis, a trajectory may be semantically likely yet clinically
unsafe (e.g., overlooking suicidal ideation) or unnecessarily expensive
(e.g., requiring many model calls).  
To mitigate such issues, we introduce two additional penalty terms:
a risk term $\mathrm{Risk}(\mathbf{ST})$ and a cost term
$\mathrm{Cost}(\mathbf{ST})$.

The overall risk of a trajectory is defined as the aggregated risk across all
sub-goals:
\begin{equation}
\mathrm{Risk}(\mathbf{ST})
= \frac{1}{n}\sum_{i=1}^{n}
\big[R_{\text{emo}}(st_i; x,g)
+ R_{\text{safety}}(st_i; x,g)\big],
\label{eq:risk_def}
\end{equation}
where $R_{\text{emo}}$ reflects diagnostic risks such as missing comorbid
emotions or producing ambiguous boundaries, while $R_{\text{safety}}$ captures
safety-related concerns such as ignoring potential self-harm cues.

The cost of a trajectory is modeled as a weighted combination of its length,
expected number of model calls, and computational latency:
\begin{equation}
\mathrm{Cost}(\mathbf{ST})
= \alpha_{\ell}\, n
+ \alpha_{\text{call}}\, C_{\text{call}}(\mathbf{ST})
+ \alpha_{\text{time}}\, C_{\text{time}}(\mathbf{ST}),
\label{eq:cost_def}
\end{equation}
where $C_{\text{call}}(\mathbf{ST})$ denotes the estimated LLM call count and
$C_{\text{time}}(\mathbf{ST})$ approximates wall-clock delay.
The coefficients $\alpha_{\ell},\alpha_{\text{call}},\alpha_{\text{time}}\ge 0$
control the trade-off among trajectory length, computational budget, and
execution time. For simplicity, we set 
$\alpha_{\ell}=\alpha_{\text{call}}=\alpha_{\text{time}}=\tfrac{1}{3}$  
in all experiments.

By combining the likelihood term with the risk and cost penalties defined in
Eqs.~\eqref{eq:risk_def}–\eqref{eq:cost_def}, the planner optimizes a
regularized objective that balances semantic plausibility, clinical safety, and computational efficiency:
\begin{align}
\pi_{\text{plan}}(g,x,p_0)
= \arg\max_{\mathbf{ST}}\;
\mathbb{E}
\Big[
&\log P
- \gamma_r\, \mathrm{Risk}
   - \gamma_c\, \mathrm{Cost}
\Big],
\label{eq:planner_main}
\end{align}
where $\gamma_r,\gamma_c\ge 0$ are high-level trade-off coefficients that
govern the planner's sensitivity to safety-related risks and computational
constraints.

This unified objective generalizes the conventional planner in
Eq.~\eqref{eq:planner_base}.  
When $\gamma_r=\gamma_c=0$, the planner collapses to pure likelihood
maximization, favoring semantically likely but potentially unsafe or
computationally expensive trajectories.  
Larger $\gamma_r$ encourages the planner to avoid unsafe or clinically risky
sub-goals, promoting conservative exploration and more reliable emotion
coverage; larger $\gamma_c$ enforces compact trajectories that reduce model
invocations and latency.

Intuitively, Eq.~\eqref{eq:planner_main} provides a structured mechanism for
controlling exploration behaviors: the likelihood term promotes semantic
coherence in the latent space; the risk penalty restricts unsafe reasoning
paths (e.g., skipping comorbid emotions or ignoring critical cues); and the cost term prevents unnecessary branching or excessive refinement.  
This yields trajectories that are not only linguistically and semantically
sound, but also clinically reliable and computationally efficient, ensuring
stable downstream refinement by the multi-agent system.

\textbf{Planning procedure.}
Operationally, APOLO implements this objective in three steps:
\begin{enumerate}
    \item 
    For each $(g,x)$, sample or approximate $z\sim q(z\mid g,x)$ using the
    underlying language model, capturing task semantics and risk signals.
    \item 
    Conditioned on $(z,p_0)$, generate a small set of candidate trajectories
    $\{\mathbf{ST}^{(k)}\}_{k=1}^{K}$ using the Planner’s policy 
    $P(\mathbf{ST}\mid z,p_0)$.
    \item 
    For each candidate, compute
    \(
    U(\mathbf{ST}^{(k)})
   ,
    \)
    as in Eq.~\eqref{eq:planner_main} and select
    \(
    \mathbf{ST}^* = \arg\max_{k} U(\mathbf{ST}^{(k)}).
    \)
\end{enumerate}

The selected trajectory $\mathbf{ST}^*$ is then passed to the
Teacher–Critic–Student triad as a high-level optimization path.
In this way, the overall refinement process is guided by a planner that is
explicitly aware of both clinical risk and computational cost, rather than
optimizing prompts in an unconstrained or purely heuristic manner.


\subsection{Joint Policy Optimization}
\label{sec:interaction}

Given the risk-aware and cost-constrained trajectory $\mathbf{ST}$ generated by 
the Planner, APOLO performs structured multi-step refinement through a 
\textit{Teacher–Critic–Student} triad. 
This refinement is essential for linguistic emotion diagnosis, where emotional 
expressions are often implicit, multi-label, and safety-sensitive. 
The triad collaboratively resolves ambiguity, strengthens diagnostic reasoning, 
and produces increasingly consistent and clinically aligned prompts.

At refinement step $i$, the Teacher proposes an emotion-aware probing query 
$q_i$ based on the sub-goal $st_i$ and the previous prompt $p_{i-1}$. 
The Critic evaluates the proposal along three dimensions: clarity, diagnostic 
relevance, and safety sensitivity, and returns structured feedback $c_i$. 
The Student incorporates the pair $(q_i,c_i)$ and updates the working prompt 
$p_i$.  
This interaction process follows  
$q_i = \pi_{\text{t}}(st_i,p_{i-1})$,  
$c_i = \pi_{\text{c}}(q_i)$,  
$p_i = \pi_{\text{s}}((q_i,c_i),p_{i-1})$,  
and a latent-state transition  
$s_i \!\sim\! \mathcal{T}(s_{i-1},(q_i,c_i))$,  
with $o_i = p_i$.

Because clinical emotion expression frequently spans multiple interacting 
symptoms, such as depression co-occurring with anxiety or guilt, isolated 
refinement steps may lose semantic continuity.  
To maintain coherence, each agent conditions not only on $(st_i,p_{i-1})$ but 
also on the historical interaction sequence  
\begin{equation}
\mathcal{H}_{<i} = \{(q_j,c_j,p_j)\}_{j<i}.
\end{equation}
The corresponding context-augmented update is:
\begin{equation}
\begin{aligned}
q_i &= \pi_{\text{t}}(st_i, p_{i-1}, \mathcal{H}_{<i}), \\
c_i &= \pi_{\text{c}}(q_i, \mathcal{H}_{<i}), \\
p_i &= \pi_{\text{s}}((q_i,c_i), p_{i-1}, \mathcal{H}_{<i}).
\end{aligned}
\label{eq:socratic_context}
\end{equation}
This formulation aligns with the POMDP transition dynamics: each refinement step 
updates the latent dialogue state and progressively reduces uncertainty across 
interdependent emotional dimensions.

\textbf{Joint policy optimization.}
Collectively, the Teacher, Critic, and Student form a joint policy 
$\Pi = \{\pi_{\text{t}},\,\pi_{\text{c}},\,\pi_{\text{s}}\}$ operating in the 
latent state space.  
APOLO learns $\Pi$ to maximize diagnostic performance while ensuring 
consistency with the planner’s trajectory.  
The optimization objective is:
\begin{equation}
\max_{\Pi}\;
\mathbb{E}_{(x,Y)\sim D}
\left[
\mathcal{R}(\Pi)
- \lambda
\sum_{i=1}^{n}
\mathcal{L}_{\text{align}}\big((q_i,c_i),\, st_i\big)
\right],
\label{eq:joint_policy}
\end{equation}
where $\mathcal{R}(\Pi)$ measures multi-label diagnostic accuracy and safety 
compliance, and $\mathcal{L}_{\text{align}}$ penalizes deviations from the 
sub-goals planned in Section~\ref{sec:plan}.  
The coefficient $\lambda$ controls the balance between adaptive exploration and 
goal-consistent refinement.

\subsection{Evaluation and Iterative Convergence}
\label{sec:feedback}

After completing the refinement steps, the final prompt 
$p_\ell = p_n$ is evaluated by the \textit{Target} agent on the 
held-out test set. 
The evaluation provides an external performance signal that measures how well 
the entire Planner–Teacher–Critic–Student pipeline improves diagnostic 
accuracy. 
Formally, at outer iteration $t$ the reward is:
\begin{equation}
\mathcal{R}^{(t)} =
\sum_{(x,Y)\in D_{\text{test}}}
f\big(\pi_{\text{tar}}(x;\, p_\ell^{(t)}),\, Y\big),
\end{equation}
where $f(\cdot)$ denotes a task-specific metric 
(e.g., micro-F1 or Jaccard for multi-label emotion diagnosis).

To ensure computational efficiency and prevent unnecessary refinement,  
APOLO employs an adaptive early-stopping rule based on marginal reward gain:  
\begin{equation}
\Delta \mathcal{R}^{(t)}
= \mathcal{R}^{(t)} - \mathcal{R}^{(t-1)}.
\end{equation}
The optimization halts when
\[
\Delta \mathcal{R}^{(t)} \le \delta 
\quad\text{or}\quad t \ge I,
\]
where $\delta$ is a minimum improvement threshold 
and $I$ is the maximum number of outer iterations.

This evaluation–iteration mechanism closes the optimization loop, ensuring
that APOLO converges to a prompt that achieves strong diagnostic performance
without excessive computation or over-refinement.


\subsection{Theoretical Insights}
\label{sec:theory}

The performance gain achieved by APOLO integrates two complementary factors:
(i) the risk-aware and cost-constrained planner, which restricts unsafe or 
inefficient exploration paths, and 
(ii) the multi-agent Socratic refinement mechanism, which provides structured 
pseudo-gradient updates to progressively improve prompt quality.  
Together, these components produce stable and interpretable optimization 
dynamics within the POMDP framework.

\textbf{Unified improvement bound.}
Let $(q_i,c_i)$ denote the Teacher–Critic action at refinement step $i$, 
with expected advantage $A_i$ and bounded variance $\sigma^2$.  
Let $B_r$ and $B_c$ provide upper bounds for the cumulative risk and cost 
penalties incurred along the planned trajectory.  
Then, after $n$ refinement steps, the expected improvement satisfies:
\begin{equation}
\mathbb{E}[\mathcal{R}(p_n)] - \mathcal{R}(p_0)
\;\ge\;
\sum_{i=1}^{n}
\left(
A_i - \frac{\sigma^2}{2\lambda}
\right)
- \gamma_r B_r
- \gamma_c B_c.
\label{eq:improvement_bound}
\end{equation}
The first term reflects progressive improvement contributed by 
Socratic pseudo-gradients, while the last two terms account for 
explicit penalties from the risk- and cost-aware planning objective.  
Equation~\eqref{eq:improvement_bound} shows that APOLO improves expected 
reward monotonically up to explicit, bounded offsets introduced for safety 
and computational efficiency.

\textit{Proof sketch.}
We view each refinement step $i$ as a KL-regularized policy update for the
joint policy $\Pi$. 
Standard analysis of KL-constrained policy improvement shows that the expected
reward gain at step $i$ satisfies
\[
\Delta \mathcal{R}_i 
= \mathbb{E}[\mathcal{R}(p_i)] - \mathbb{E}[\mathcal{R}(p_{i-1})]
\;\ge\;
A_i - \frac{\sigma^2}{2\lambda},
\]
where $A_i$ is the expected advantage of taking action $(q_i,c_i)$ under the
current policy and $\sigma^2$ bounds the variance of the corresponding return.
Summing over $i=1,\dots,n$ yields
\[
\mathbb{E}[\mathcal{R}(p_n)] - \mathcal{R}(p_0)
= \sum_{i=1}^{n} \Delta \mathcal{R}_i
\;\ge\;
\sum_{i=1}^{n}
\left(
A_i - \frac{\sigma^2}{2\lambda}
\right).
\]

The planner further introduces explicit penalties on risk and cost along the
trajectory. 
By assumption, the accumulated risk and cost terms are bounded by $B_r$ and
$B_c$, and the corresponding contributions to the objective are scaled by
$\gamma_r$ and $\gamma_c$, respectively. 
Therefore, incorporating these penalties subtracts at most 
$\gamma_r B_r + \gamma_c B_c$ from the achievable improvement, which leads
directly to the bound in Eq.~\eqref{eq:improvement_bound}. \hfill$\square$

The complete optimization procedure is summarized in Algorithm~\ref{alg:training}.

\begin{algorithm}[t]
\footnotesize
\caption{APOLO Optimization Procedure}
\label{alg:training}
\begin{algorithmic}[1]
\State \textbf{Input:} Dataset $\mathcal{D}$, initial prompt $p_0$, 
thresholds $(\delta,\gamma_{\text{risk}},\gamma_{\text{cost}})$, 
maximum iterations $I$
\State \textbf{Output:} Optimized prompt $p^*$

\State \textit{Planner}: Generate risk-aware and cost-constrained trajectory 
$\mathbf{ST}=\{st_1,\dots,st_n\}$
\State Initialize $p_0^{(1)} \gets p_0$, \; $\mathcal{R}^{(0)} \gets 0$

\For{$t=1$ to $I$}
    \For{$i=1$ to $n$}  \hfill\textit{// Socratic refinement}
        \State $q_i \gets \pi_{\text{t}}(st_i,\, p_{i-1}^{(t)})$
        \Repeat
            \State $c_i \gets \pi_{\text{c}}(q_i)$  \hfill (clarity, relevance, safety)
            \State Teacher revises $q_i$ if necessary
        \Until{quality and safety are satisfied}
        \State $p_i^{(t)} \gets \pi_{\text{s}}((q_i,c_i),\, p_{i-1}^{(t)})$
    \EndFor

    \State $p_\ell^{(t)} \gets p_n^{(t)}$ \hfill\textit{// Final prompt}
    \State $\mathcal{R}^{(t)}
        = \sum_{(x,Y)\in D_{\text{test}}}
            f(\pi_{\text{tar}}(x;\, p_\ell^{(t)}),\, Y)$

    \If{$\mathcal{R}^{(t)} - \mathcal{R}^{(t-1)} \le \delta$}
        \State \textbf{break}
    \EndIf
\EndFor

\State \textbf{return} $p^* \gets p_\ell^{(t)}$
\end{algorithmic}
\end{algorithm}

\section{Experiments \label{sec:experiments}}
In this section, we conduct comprehensive experiments to evaluate the proposed APOLO framework. Section~\ref{sec:setup} introduces the experimental setup, including datasets, implementation details, evaluation metrics, and baselines. Section~\ref{sec:main_results} presents the main results, demonstrating APOLO’s superior performance across diverse tasks and models. Section~\ref{sec:efficiency} further analyzes its computational efficiency, showing that APOLO is both effective and practical.

\subsection{Experiment Setup}
\label{sec:setup}

\begin{table*}[t]
\caption{Comprehensive comparison of APOLO and baseline methods on six emotion diagnosis benchmarks. 
For single-label datasets, we report both Macro F1(M-F1) and Micro F1(m-F1) scores (\%). 
For the multi-label \textit{DepressionEmo} dataset, we additionally present Exact Match Ratio (EMR, \%) and Partial Match Accuracy (PMA, \%). 
The best and second-best results are highlighted in \textbf{bold} and \underline{underlined}, respectively. }
\label{tab:main_results}
\centering
\renewcommand{\arraystretch}{1.2}
\resizebox{\textwidth}{!}{
\begin{tabular}{@{}c|
cc cc cc cc cc|
cccc|
cc@{}}
\toprule
\multirow{2}{*}{\textbf{Method}} 
& \multicolumn{2}{c}{\textbf{DailyDialog}}
& \multicolumn{2}{c}{\textbf{EmoryNLP}}
& \multicolumn{2}{c}{\textbf{PELD}}
& \multicolumn{2}{c}{\textbf{RECCON}}
& \multicolumn{2}{c|}{\textbf{EmotionX}}
& \multicolumn{4}{c|}{\textbf{DepressionEmo}}
& \multicolumn{2}{c@{}}{\textbf{Avg.}} \\ 
\cmidrule(lr){2-11} \cmidrule(lr){12-15} \cmidrule(lr){16-17}
& \textbf{M-F1} & \textbf{m-F1}
& \textbf{M-F1} & \textbf{m-F1}
& \textbf{M-F1} & \textbf{m-F1}
& \textbf{M-F1} & \textbf{m-F1}
& \textbf{M-F1} & \textbf{m-F1}
& \textbf{M-F1} & \textbf{m-F1} & \textbf{EMR} & \textbf{PMA}
& \textbf{M-F1} & \textbf{m-F1} \\
\midrule
\multicolumn{17}{c}{\cellcolor{gray!20}\textbf{GPT-5-mini}} \\
\midrule
Origin       & 27.37 & 75.26 & 41.51 & 46.16 & 46.09 & 60.73 & 37.71 & 43.01 & 32.84 & 46.38 & 58.26 & 69.32 & 11.15 & 58.61 & 40.63 & 56.81 \\
CoT(ZS)      & 28.15 & 75.98 & 42.33 & 47.06 & 46.95 & 61.87 & 38.54 & 44.29 & 33.67 & 47.52 & 59.98 & 70.97 & 13.36 & 62.36 & 41.60 & 57.95 \\
CoT(FS)      & 29.02 & 76.85 & 43.10 & 48.19 & 47.88 & 63.00 & 39.42 & 45.65 & 34.51 & 48.79 & 61.85 & 73.07 & 16.00 & 68.65 & 42.63 & 59.26 \\
APE          & 29.56 & 77.42 & 43.65 & 48.95 & 48.45 & 63.78 & 39.99 & 46.43 & 35.02 & 49.54 & 63.01 & 74.50 & 18.10 & 73.40 & 43.28 & 60.10 \\
ProTeGi      & 30.11 & 77.98 & 44.21 & 49.70 & 49.03 & 64.55 & 40.57 & 47.21 & 35.54 & 50.33 & 64.22 & 75.94 & 20.42 & 78.15 & 43.95 & 60.95 \\
OPRO         & \underline{31.25} & \underline{79.12} & 45.15 & 50.83 & \underline{50.23} & \underline{66.00} & \underline{41.77} & \underline{48.75} & \underline{36.63} & \underline{51.89} & \underline{66.45} & \underline{78.70} & 24.28 & \underline{86.75} & \underline{45.25} & \underline{62.55} \\
PE2          & 30.78 & 78.68 & \underline{45.42} & \underline{51.20} & 49.75 & 65.45 & 41.28 & 48.12 & 36.18 & 51.27 & 65.59 & 77.48 & \underline{24.61} & 83.00 & 44.83 & 62.03 \\
\midrule
\rowcolor{gray!10}
\textbf{APOLO (Ours)} & \textbf{34.11} & \textbf{82.34} & \textbf{48.01} & \textbf{54.22} & \textbf{53.15} & \textbf{69.23} & \textbf{44.52} & \textbf{51.89} & \textbf{39.21} & \textbf{55.02} & \textbf{70.11} & \textbf{82.45} & \textbf{28.92} & \textbf{91.17} & \textbf{48.19} & \textbf{65.86} \\

\midrule

\multicolumn{17}{c}{\cellcolor{gray!20}\textbf{DeepSeek-V3}} \\
\midrule
Origin       & 23.47 & 36.16 & 39.88 & 43.15 & 45.54 & 53.83 & 45.92 & 47.82 & 31.81 & 36.64 & 65.04 & 75.06 & 10.82 & 86.31 & 41.94 & 48.78 \\
CoT(ZS)      & 24.55 & 39.87 & 40.76 & 44.95 & 46.63 & 55.02 & 46.91 & 49.33 & 32.75 & 38.01 & 66.82 & 76.93 & 12.91 & 88.08 & 43.07 & 50.69 \\
CoT(FS)      & 25.68 & 43.71 & 41.67 & 46.84 & 47.75 & 56.24 & 47.95 & 50.91 & 33.72 & 39.41 & 68.75 & 79.03 & 15.23 & 89.96 & 44.25 & 52.69 \\
APE          & 26.34 & 46.21 & 42.24 & 47.89 & 48.43 & 57.01 & 48.60 & 51.88 & 34.33 & \underline{43.09} & 70.01 & 80.46 & 17.11 & 90.95 & 44.99 & 54.42 \\
ProTeGi      & 27.01 & 48.81 & 42.82 & 48.95 & 49.12 & 57.80 & 49.26 & 52.87 & 34.95 & 41.24 & 71.30 & 81.90 & 19.21 & 92.05 & 45.74 & 55.26 \\
OPRO         & \underline{28.31} & \underline{53.19} & \underline{43.95} & \underline{50.83} & 50.35 & 59.13 & \underline{50.59} & \underline{54.72} & \underline{36.21} & 42.96 & \underline{73.69} & \underline{84.66} & \underline{22.96} & \underline{94.04} & \underline{47.18} & \underline{57.58} \\
PE2          & 27.82 & 51.55 & 43.51 & 50.08 & \underline{50.55} & \underline{59.41} & 50.05 & 53.99 & 35.69 & 42.28 & 72.76 & 83.55 & 21.41 & 93.38 & 46.73 & 56.81 \\
\midrule
\rowcolor{gray!10}
\textbf{APOLO (Ours)} & \textbf{31.02} & \textbf{57.21} & \textbf{46.55} & \textbf{54.89} & \textbf{53.48} & \textbf{62.50} & \textbf{53.61} & \textbf{58.03} & \textbf{38.84} & \textbf{45.99} & \textbf{77.23} & \textbf{88.08} & \textbf{27.15} & \textbf{96.80} & \textbf{50.12} & \textbf{61.12} \\
\midrule

\multicolumn{17}{c}{\cellcolor{gray!20}\textbf{Qwen3-32B}} \\
\midrule
Origin       & 28.11 & 76.01 & 42.43 & 47.21 & 47.02 & 61.86 & 38.65 & 44.13 & 33.76 & 47.49 & 59.33 & 70.42 & 11.92 & 60.04 & 41.55 & 57.85 \\
CoT(ZS)      & 29.02 & 76.89 & 43.29 & 48.12 & 47.99 & 63.02 & 39.51 & 45.44 & 34.62 & 48.66 & 61.11 & 72.19 & 14.24 & 63.91 & 42.59 & 59.05 \\
CoT(FS)      & 29.98 & 77.80 & 44.18 & 49.25 & 48.98 & 64.24 & 40.45 & 46.85 & 35.51 & 49.93 & 63.05 & 74.28 & 16.89 & 70.42 & 43.69 & 60.39 \\
APE          & 30.55 & 78.40 & 44.76 & 50.00 & 49.58 & 65.03 & 41.05 & 47.67 & 36.05 & 50.75 & 64.25 & 75.83 & 19.21 & 75.28 & 44.37 & 61.28 \\
ProTeGi      & 31.13 & 79.01 & 45.35 & 50.83 & 50.19 & 65.84 & 41.66 & 48.48 & 36.60 & 51.56 & 65.49 & 77.26 & 21.41 & 80.13 & 45.07 & 62.16 \\
OPRO          & \underline{32.33} & \underline{80.21} & \underline{46.52} & \underline{52.33} & \underline{51.46} & \underline{67.37} & 42.71 & 49.88 & \underline{37.75} & \underline{53.19} & \underline{67.82} & \underline{80.02} & \underline{25.61} & 89.85 & \underline{46.43} & \underline{63.83} \\
PE2         & 31.84 & 79.75 & 46.06 & 51.73 & 50.94 & 66.76 & \underline{42.95} & \underline{50.12} & 37.28 & 52.54 & 66.91 & 78.92 & 23.95 & \underline{90.18} & 46.00 & 63.30 \\
\midrule
\rowcolor{gray!10}
\textbf{APOLO (Ours)} & \textbf{35.24} & \textbf{83.51} & \textbf{49.21} & \textbf{55.50} & \textbf{54.41} & \textbf{70.62} & \textbf{45.72} & \textbf{53.18} & \textbf{40.42} & \textbf{56.42} & \textbf{71.55} & \textbf{83.89} & \textbf{30.02} & \textbf{93.38} & \textbf{49.43} & \textbf{67.19} \\
\bottomrule
\end{tabular}}

\end{table*}

\textbf{Datasets.}
To ensure a comprehensive and rigorous evaluation of APOLO's capabilities, our experiments are conducted on a carefully curated selection of six public datasets. This collection spans a wide spectrum of conversational emotion analysis tasks, systematically testing the framework's adaptability and effectiveness. We include foundational multi-turn dialogue datasets like DailyDialog~\cite{li2017dailydialog} for general conversational scenarios and EmoryNLP~\cite{zahiri2018emorynlp} to challenge the model with longer contextual dependencies. To assess the framework's ability to handle more complex inference, we incorporate PELD~\cite{peld} and RECCON~\cite{poria2021recognizing}, both of which introduce the task of identifying emotion-cause pairs. Furthermore, to test the robustness of our method in noisy, real-world settings, we utilize the EmotionX~\cite{shmueli2019socialnlpemotionx2019challenge} dataset, composed of multi-party dialogues from diverse social media platforms. Finally, to evaluate performance on specialized, fine-grained tasks, we include DepressionEmo~\cite{rahman2024depressionemo}, which uniquely presents the challenge of multi-label classification of psychological distress signals. Collectively, this diverse suite of datasets allows us to validate the generalizability of APOLO across various complexities, domains, and annotation schemes inherent in linguistic emotion diagnosis.

\textbf{Implementation Details.}
To demonstrate its versatility and robustness, we conduct parallel experiments utilizing two distinct large language models as the backbone for all agents: \texttt{GPT-5-mini}~\cite{openai_gpt5}, \texttt{DeepSeek-V3}~\cite{liu2024deepseek} and \texttt{Qwen-32B}~\cite{yang2025qwen3}. Within any single experimental run, all agents (\textit{Planner}, \textit{Teacher}, \textit{Student}, \textit{Critic}, and \textit{Target}) are powered by the same backbone model to ensure internal consistency. For all generative inferences, the decoding temperature is uniformly set to $0.6$ to balance creativity and factual coherence.

The automated optimization process is configured for a maximum of $I=10$ iterations. To enhance efficiency and prevent overfitting on minor gains, an early stopping mechanism is implemented. The process terminates if the accuracy improvement on the validation set between two consecutive iterations falls below a threshold of $\delta = 0.01$. Furthermore, to streamline the optimization path, each \textit{assess-adjust} cycle within an iteration is constrained to a single revision per step. Following each optimization cycle, the performance of the generated prompt is rigorously evaluated by the \textit{Target} agent on the entire test set of the respective dataset to track the optimization trajectory comprehensively.

\textbf{Evaluation Metrics.}
The selection of evaluation metrics is carefully aligned with the specific characteristics of each task to ensure a fair and comprehensive assessment of model performance. For the multi-class classification tasks, which constitute the majority of our datasets (e.g.,~DailyDialog, EmoryNLP), the \textbf{Macro F1-score} is adopted as the primary evaluation metric. This measure is particularly suitable for emotion recognition as it alleviates potential bias from class imbalance by computing the F1-score independently for each emotion category and then averaging them. Formally, for a set of classes $C$, where $\text{F1}_c$ is the F1-score for class $c$:
\begin{equation}
    \text{Macro F1} = \frac{1}{|C|} \sum_{c \in C} \text{F1}_c
    \label{eq:macrof1}
\end{equation}
This approach assigns equal importance to both frequent and rare emotions. To provide a complementary perspective on overall performance, we also report the \textbf{Micro F1-score}, which aggregates the contributions of all samples to compute a single metric. It is calculated from the global sum of true positives ($\text{TP}_{\mu}$), false positives ($\text{FP}_{\mu}$), and false negatives ($\text{FN}_{\mu}$):
\begin{equation}
    \text{Micro F1} = \frac{2 \cdot \text{TP}_{\mu}}{2 \cdot \text{TP}_{\mu} + \text{FP}_{\mu} + \text{FN}_{\mu}}
    \label{eq:microf1}
\end{equation}

For the multi-label classification task in the DepressionEmo dataset, where each sample may be associated with multiple emotional states, a multi-faceted evaluation strategy was employed. For a dataset of $N$ samples, let $Y_i$ be the set of true labels and $Z_i$ be the set of predicted labels for the $i$-th sample. The \textbf{Exact Match Ratio (EMR)} provides a strict measure by quantifying the percentage of instances whose predicted label sets exactly match the ground truth:
\begin{equation}
    \text{EMR} = \frac{1}{N} \sum_{i=1}^{N} \mathbb{I}(Y_i = Z_i)
    \label{eq:emr}
\end{equation}
where $\mathbb{I}(\cdot)$ is the indicator function. To complement this stringent criterion and account for partially correct predictions, the \textbf{Partial Match Accuracy (PMA)} was introduced, which deems a prediction correct if the model successfully identifies more than half of the true labels:
\begin{equation}
    \text{PMA} = \frac{1}{N} \sum_{i=1}^{N} \mathbb{I}\left(|Y_i \cap Z_i| > \frac{|Y_i|}{2}\right)
    \label{eq:pma}
\end{equation}

Furthermore, to ensure a comprehensive assessment of per-label performance, we also report both the label-based Macro F1-score (calculated as in Eq.~\ref{eq:macrof1}) for a balanced view of performance across all labels, and the label-based Micro F1-score (calculated as in Eq.~\ref{eq:microf1}) for an aggregate measure of correctness. Collectively, this comprehensive suite of metrics provides a robust and nuanced evaluation framework for prompt performance across both single-label and multi-label diagnostic scenarios.

\textbf{Baselines.}
We compare APOLO against three categories of baselines: original prompts, CoT prompts, and several state-of-the-art(SOTA) APO approaches.

(1) Original prompts refer to the default task instructions provided within each dataset, which typically contain minimal task-specific guidance and therefore serve as a straightforward reference point for performance evaluation.

(2) For the CoT (Zero-Shot) baseline, we prepend the phrase “Let’s think step by step” to each task instruction to encourage explicit intermediate reasoning. Building upon this, the CoT (Few-Shot) baseline incorporates an additional in-context example that illustrates the reasoning process, thereby enhancing the model’s ability to generalize from limited demonstrations.

(3) In addition, we include several representative APO methods from recent literature, namely APE~\cite{zhou2023large}, Prompt Optimization with Textual Gradients (ProTeGi)~\cite{pryzant2023automatic}, Optimization by PROmpting (OPRO)~\cite{Yang0LLLZC24}, and Prompt Engineer 2 (PE2)~\cite{ye2023prompt}.
APE formulates prompt design as a search problem, where a LLM autonomously generates, evaluates, and selects optimal prompt candidates based on self-assessment. ProTeGi refines prompts through gradient-inspired textual feedback derived from model prediction errors, enabling more directed optimization. OPRO conceptualizes prompt optimization as an iterative meta-learning process, in which the LLM itself functions as an optimizer that continually improves prompt formulations. Building upon this idea, PE2 introduces a structured meta-prompting framework that performs multi-round refinement, yielding more robust and generalizable prompts across diverse tasks.

\subsection{Main Results}
\label{sec:main_results}
\textbf{APOLO achieves an overall improvement across all benchmarks, setting a new SOTA in APO for emotion diagnosis.}
Across six benchmark datasets, as summarized in Table~\ref{tab:main_results}, APOLO achieves the highest overall performance with an average Macro F1 of \textbf{49.25\%} and Micro F1 of \textbf{64.72\%}, surpassing the best baseline (OPRO) by \textbf{2.96\%} and \textbf{3.40\%}, respectively.
The improvement is not limited to aggregated scores: APOLO also increases the Exact Match Ratio (EMR) to \textbf{28.70\%} and the Partial Match Accuracy (PMA) to \textbf{93.78\%}, indicating stronger consistency and precision in handling complex, multi-label emotional data.
Together, these results demonstrate that APOLO’s optimization framework substantially enhances both the accuracy and reliability of emotion understanding across diverse diagnostic scenarios.


\textbf{APOLO delivers consistent and substantial improvements across both single-label and multi-label emotion diagnosis tasks.}
For single-label datasets such as \textit{DailyDialog} and \textit{EmotionX}, APOLO achieves remarkable gains in Macro F1, suggesting its superior capability in identifying minority or subtle emotional categories that are often underrepresented during training.
This improvement reflects APOLO’s enhanced ability to capture fine-grained affective nuances and mitigate the effects of class imbalance.
In multi-label settings, particularly on the \textit{DepressionEmo} dataset, APOLO substantially boosts the Exact Match Ratio (EMR) and Partial Match Accuracy (PMA), with increases of up to \textbf{+4.41\%} and \textbf{+3.57\%}, respectively.
These gains indicate that APOLO more effectively models complex emotional co-occurrences and overlapping affective states, such as anxiety coexisting with fatigue or sadness.
Moreover, the consistently larger relative improvement in Macro F1 over Micro F1 demonstrates that APOLO not only enhances overall predictive performance but also strengthens its sensitivity to infrequent yet clinically meaningful emotions.

\textbf{APOLO demonstrates consistent advantages across heterogeneous foundational models, reflecting its robustness, adaptability, and strong generalization capability.}
When applied to three representative LLM backbones, namely \texttt{GPT-5-mini} (closed-source), \texttt{DeepSeek-V3}, and \texttt{Qwen3-32B} (both open-source), APOLO consistently achieves superior results under all experimental configurations.
Across these models, it delivers average Macro F1 improvements of \textbf{+2.94\%}, \textbf{+2.94\%}, and \textbf{+3.00\%} over the strongest baseline, respectively, together with comparable gains in Micro F1, indicating stable and comprehensive performance enhancement.
Notably, APOLO’s improvements generalize well beyond specific architectures or vendor ecosystems. It achieves robust gains on both proprietary systems such as GPT-5 and open-source counterparts including DeepSeek and Qwen, suggesting that its Socratic optimization mechanism effectively enhances emotional reasoning independent of model origin, architecture, or training pipeline.
Moreover, the relative performance hierarchy among baseline methods remains consistent across all backbones, reinforcing the stability and reproducibility of APOLO’s optimization dynamics.

\begin{figure}[t]
	\large
	\centering
	\includegraphics[width=\columnwidth]{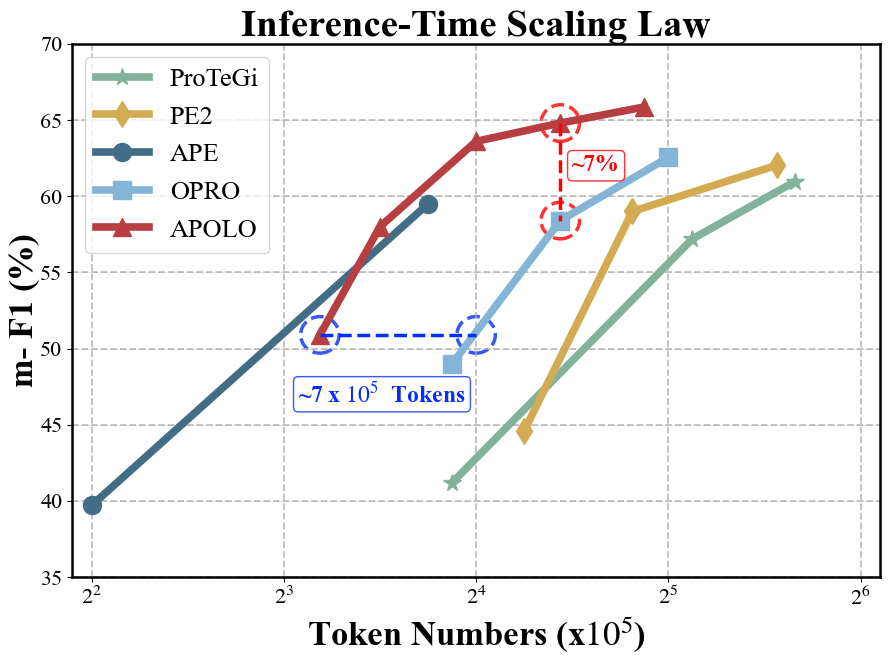}
	\caption{
    Inference-time scaling law for APOLO and baseline methods, evaluated on datasets related to medical emotion diagnosis, illustrating the relationship between Micro F1-score and total token consumption during optimization.}
	\label{scaling}
\end{figure}

\subsection{Efficiency Analysis}
\label{sec:efficiency}
\textbf{APOLO Consistently Achieves the Highest Computational Efficiency for emotion diagnosis.}
The balance between resource consumption and performance improvement serves as a fundamental criterion for evaluating the practical effectiveness of APO systems in emotion diagnosis~\cite{yang2024harnessing}.
As illustrated in Figure~\ref{scaling}, APOLO consistently surpasses all baseline methods in terms of computational efficiency, demonstrating a more favorable inference-time scaling trend across a wide range of APO tasks related to emotion diagnosis. This indicates that APOLO not only achieves faster convergence but also maintains stable performance even as task complexity or model scale increases.

As shown in Figure~\ref{scaling}, the red APOLO curve is located in the upper-left region of the plot, indicating its superior efficiency in achieving higher accuracy with fewer generated tokens.
Specifically, when the number of generated tokens is fixed at approximately $2^{4.4} \times 10^5$, APOLO achieves about 7\% higher average Micro F1-score than OPRO on emotion diagnosis benchmarks (as highlighted by the red dashed annotation).
Conversely, to reach the same performance level as APOLO, OPRO must generate roughly $7 \times 10^5$ more tokens (as indicated by the blue dashed arrow), further emphasizing APOLO’s advantage in balancing performance and computational cost.
In contrast, competing methods generally require a significantly larger number of generated tokens to reach comparable results, which implies greater computational overhead and less efficient utilization of inference resources under emotion diagnosis settings.

These findings clearly highlight APOLO’s superior capability to balance effectiveness and efficiency through its structured optimization design.
By first conducting high-level task decomposition and then applying step-wise Socratic refinement tailored to emotion diagnosis reasoning patterns, APOLO facilitates hierarchical reasoning and adaptive resource allocation, allowing the model to focus computational effort where it is most impactful.
Such a strategy substantially reduces redundant generation, mitigates unnecessary reasoning steps, and ultimately ensures both robust performance gains and sustainable computational cost throughout the entire APO pipeline.


\begin{table*}[t]
\caption{Ablation study of APOLO's key components, conducted on medical emotion diagnosis tasks using three distinct backbone models. We systematically remove the Planner (${w/o}_\text{Plan}$), the entire Socratic module (${w/o}_\text{Soc}$), and the Critic (${w/o}_\text{Cri}$). The performance drop ($\Delta$) relative to the full APOLO framework is shown in \textcolor{ForestGreen}{green}.}
\label{tab:ablation}
\centering
\renewcommand{\arraystretch}{1.2}
\resizebox{\textwidth}{!}{
\begin{tabular}{@{}c|
cc cc cc cc cc|
cccc|
cc@{}}
\toprule
\multirow{2}{*}{\textbf{Method}} 
& \multicolumn{2}{c}{\textbf{DailyDialog}}
& \multicolumn{2}{c}{\textbf{EmoryNLP}}
& \multicolumn{2}{c}{\textbf{PELD}}
& \multicolumn{2}{c}{\textbf{RECCON}}
& \multicolumn{2}{c|}{\textbf{EmotionX}}
& \multicolumn{4}{c|}{\textbf{DepressionEmo}}
& \multicolumn{2}{c@{}}{\textbf{Avg.}} \\ 
\cmidrule(lr){2-11} \cmidrule(lr){12-15} \cmidrule(lr){16-17}
& \textbf{M-F1} & \textbf{m-F1}
& \textbf{M-F1} & \textbf{m-F1}
& \textbf{M-F1} & \textbf{m-F1}
& \textbf{M-F1} & \textbf{m-F1}
& \textbf{M-F1} & \textbf{m-F1}
& \textbf{M-F1} & \textbf{m-F1} & \textbf{EMR} & \textbf{PMA}
& \textbf{M-F1} & \textbf{m-F1} \\
\midrule
\multicolumn{17}{c}{\cellcolor{gray!20}\textbf{GPT-5-mini}} \\
\midrule
\rowcolor{gray!10}
\textbf{APOLO (Ours)} & \textbf{34.11} & \textbf{82.34} & \textbf{48.01} & \textbf{54.22} & \textbf{53.15} & \textbf{69.23} & \textbf{44.52} & \textbf{51.89} & \textbf{39.21} & \textbf{55.02} & \textbf{70.11} & \textbf{82.45} & \textbf{28.92} & \textbf{91.17} & \textbf{48.19} & \textbf{65.86} \\
\midrule
${w/o}_\text{Plan}$ & 31.18 & 79.01 & 44.53 & 50.08 & 49.25 & 65.18 & 41.05 & 48.02 & 36.15 & 51.34 & 64.98 & 78.15 & 23.62 & 84.99 & 44.52 & 61.96 \\
\textcolor{ForestGreen}{$\quad\Delta$} & \textcolor{ForestGreen}{-2.93} & \textcolor{ForestGreen}{-3.33} & \textcolor{ForestGreen}{-3.48} & \textcolor{ForestGreen}{-4.14} & \textcolor{ForestGreen}{-3.90} & \textcolor{ForestGreen}{-4.05} & \textcolor{ForestGreen}{-3.47} & \textcolor{ForestGreen}{-3.87} & \textcolor{ForestGreen}{-3.06} & \textcolor{ForestGreen}{-3.68} & \textcolor{ForestGreen}{-5.13} & \textcolor{ForestGreen}{-4.30} & \textcolor{ForestGreen}{-5.30} & \textcolor{ForestGreen}{-6.18} & \textcolor{ForestGreen}{-3.66} & \textcolor{ForestGreen}{-3.90} \\
\midrule

${w/o}_\text{Soc}$ & 28.55 & 75.88 & 40.91 & 46.54 & 45.81 & 61.98 & 38.12 & 44.25 & 33.04 & 48.18 & 60.15 & 73.07 & 18.87 & 79.25 & 41.10 & 58.32 \\
\textcolor{ForestGreen}{$\quad\Delta$} & \textcolor{ForestGreen}{-5.56} & \textcolor{ForestGreen}{-6.46} & \textcolor{ForestGreen}{-7.10} & \textcolor{ForestGreen}{-7.68} & \textcolor{ForestGreen}{-7.34} & \textcolor{ForestGreen}{-7.25} & \textcolor{ForestGreen}{-6.40} & \textcolor{ForestGreen}{-7.64} & \textcolor{ForestGreen}{-6.17} & \textcolor{ForestGreen}{-6.84} & \textcolor{ForestGreen}{-9.96} & \textcolor{ForestGreen}{-9.38} & \textcolor{ForestGreen}{-10.05} & \textcolor{ForestGreen}{-11.92} & \textcolor{ForestGreen}{-7.09} & \textcolor{ForestGreen}{-7.54} \\
\midrule

${w/o}_\text{Cri}$ & 32.54 & 80.52 & 46.18 & 52.33 & 51.08 & 67.01 & 42.88 & 50.11 & 37.55 & 53.09 & 67.53 & 80.02 & 26.05 & 88.63 & 46.29 & 63.85 \\
\textcolor{ForestGreen}{$\quad\Delta$} & \textcolor{ForestGreen}{-1.57} & \textcolor{ForestGreen}{-1.82} & \textcolor{ForestGreen}{-1.83} & \textcolor{ForestGreen}{-1.89} & \textcolor{ForestGreen}{-2.07} & \textcolor{ForestGreen}{-2.22} & \textcolor{ForestGreen}{-1.64} & \textcolor{ForestGreen}{-1.78} & \textcolor{ForestGreen}{-1.66} & \textcolor{ForestGreen}{-1.93} & \textcolor{ForestGreen}{-2.58} & \textcolor{ForestGreen}{-2.43} & \textcolor{ForestGreen}{-2.87} & \textcolor{ForestGreen}{-2.54} & \textcolor{ForestGreen}{-1.89} & \textcolor{ForestGreen}{-2.01} \\
\midrule
\multicolumn{17}{c}{\cellcolor{gray!20}\textbf{DeepSeek-V3}} \\
\midrule
\rowcolor{gray!10}
\textbf{APOLO (Ours)} & \textbf{31.02} & \textbf{57.21} & \textbf{46.55} & \textbf{54.89} & \textbf{53.48} & \textbf{62.50} & \textbf{53.61} & \textbf{58.03} & \textbf{38.84} & \textbf{45.99} & \textbf{77.23} & \textbf{88.08} & \textbf{27.15} & \textbf{96.80} & \textbf{50.12} & \textbf{61.12} \\
\midrule
${w/o}_\text{Plan}$ & 28.15 & 49.88 & 43.12 & 46.99 & 49.85 & 55.24 & 49.95 & 54.12 & 35.77 & 39.54 & 72.08 & 80.13 & 18.10 & 88.63 & 46.49 & 54.32 \\
\textcolor{ForestGreen}{$\quad\Delta$} & \textcolor{ForestGreen}{-2.87} & \textcolor{ForestGreen}{-7.33} & \textcolor{ForestGreen}{-3.43} & \textcolor{ForestGreen}{-7.90} & \textcolor{ForestGreen}{-3.63} & \textcolor{ForestGreen}{-7.26} & \textcolor{ForestGreen}{-3.66} & \textcolor{ForestGreen}{-3.91} & \textcolor{ForestGreen}{-3.07} & \textcolor{ForestGreen}{-6.45} & \textcolor{ForestGreen}{-5.15} & \textcolor{ForestGreen}{-7.95} & \textcolor{ForestGreen}{-9.05} & \textcolor{ForestGreen}{-8.17} & \textcolor{ForestGreen}{-3.64} & \textcolor{ForestGreen}{-6.80} \\
\midrule

${w/o}_\text{Soc}$ & 25.41 & 46.55 & 40.03 & 43.30 & 46.19 & 51.07 & 46.58 & 50.33 & 32.54 & 36.19 & 67.81 & 75.94 & 14.13 & 83.22 & 43.09 & 50.56 \\
\textcolor{ForestGreen}{$\quad\Delta$} & \textcolor{ForestGreen}{-5.61} & \textcolor{ForestGreen}{-10.66} & \textcolor{ForestGreen}{-6.52} & \textcolor{ForestGreen}{-11.59} & \textcolor{ForestGreen}{-7.29} & \textcolor{ForestGreen}{-11.43} & \textcolor{ForestGreen}{-7.03} & \textcolor{ForestGreen}{-7.70} & \textcolor{ForestGreen}{-6.30} & \textcolor{ForestGreen}{-9.80} & \textcolor{ForestGreen}{-9.42} & \textcolor{ForestGreen}{-12.14} & \textcolor{ForestGreen}{-13.02} & \textcolor{ForestGreen}{-13.58} & \textcolor{ForestGreen}{-7.03} & \textcolor{ForestGreen}{-10.55} \\
\midrule

${w/o}_\text{Cri}$ & 29.45 & 51.36 & 44.89 & 49.10 & 51.64 & 57.28 & 51.87 & 56.15 & 37.12 & 41.24 & 75.02 & 82.78 & 20.86 & 91.83 & 48.33 & 56.32 \\
\textcolor{ForestGreen}{$\quad\Delta$} & \textcolor{ForestGreen}{-1.57} & \textcolor{ForestGreen}{-5.85} & \textcolor{ForestGreen}{-1.66} & \textcolor{ForestGreen}{-5.79} & \textcolor{ForestGreen}{-1.84} & \textcolor{ForestGreen}{-5.22} & \textcolor{ForestGreen}{-1.74} & \textcolor{ForestGreen}{-1.88} & \textcolor{ForestGreen}{-1.72} & \textcolor{ForestGreen}{-4.75} & \textcolor{ForestGreen}{-2.21} & \textcolor{ForestGreen}{-5.30} & \textcolor{ForestGreen}{-6.29} & \textcolor{ForestGreen}{-4.97} & \textcolor{ForestGreen}{-1.79} & \textcolor{ForestGreen}{-4.80} \\

\midrule
\multicolumn{17}{c}{\cellcolor{gray!20}\textbf{Qwen-32B}} \\
\midrule
\rowcolor{gray!10}
\textbf{APOLO (Ours)} & \textbf{35.24} & \textbf{83.51} & \textbf{49.21} & \textbf{55.50} & \textbf{54.41} & \textbf{70.62} & \textbf{45.72} & \textbf{53.18} & \textbf{40.42} & \textbf{56.42} & \textbf{71.55} & \textbf{83.89} & \textbf{30.02} & \textbf{93.38} & \textbf{49.43} & \textbf{67.19} \\
\midrule
${w/o}_\text{Plan}$ & 31.21 & 79.43 & 45.23 & 51.73 & 50.02 & 66.18 & 42.38 & 49.77 & 37.14 & 52.05 & 66.81 & 79.91 & 24.39 & 87.53 & 45.47 & 63.18 \\
\textcolor{ForestGreen}{$\quad\Delta$} & \textcolor{ForestGreen}{-4.03} & \textcolor{ForestGreen}{-4.08} & \textcolor{ForestGreen}{-3.98} & \textcolor{ForestGreen}{-3.77} & \textcolor{ForestGreen}{-4.39} & \textcolor{ForestGreen}{-4.44} & \textcolor{ForestGreen}{-3.34} & \textcolor{ForestGreen}{-3.41} & \textcolor{ForestGreen}{-3.28} & \textcolor{ForestGreen}{-4.37} & \textcolor{ForestGreen}{-4.74} & \textcolor{ForestGreen}{-3.98} & \textcolor{ForestGreen}{-5.63} & \textcolor{ForestGreen}{-5.85} & \textcolor{ForestGreen}{-3.96} & \textcolor{ForestGreen}{-4.01} \\
\midrule

${w/o}_\text{Soc}$ & 28.05 & 76.12 & 41.91 & 47.29 & 46.88 & 62.02 & 38.91 & 45.43 & 33.63 & 48.63 & 62.40 & 75.72 & 19.43 & 82.67 & 41.96 & 59.20 \\
\textcolor{ForestGreen}{$\quad\Delta$} & \textcolor{ForestGreen}{-7.19} & \textcolor{ForestGreen}{-7.39} & \textcolor{ForestGreen}{-7.30} & \textcolor{ForestGreen}{-8.21} & \textcolor{ForestGreen}{-7.53} & \textcolor{ForestGreen}{-8.60} & \textcolor{ForestGreen}{-6.81} & \textcolor{ForestGreen}{-7.75} & \textcolor{ForestGreen}{-6.79} & \textcolor{ForestGreen}{-7.79} & \textcolor{ForestGreen}{-9.15} & \textcolor{ForestGreen}{-8.17} & \textcolor{ForestGreen}{-10.59} & \textcolor{ForestGreen}{-10.71} & \textcolor{ForestGreen}{-7.46} & \textcolor{ForestGreen}{-7.99} \\
\midrule

${w/o}_\text{Cri}$ & 33.14 & 81.21 & 46.85 & 53.39 & 52.04 & 68.19 & 43.72 & 51.62 & 38.42 & 53.75 & 68.52 & 81.90 & 26.93 & 89.74 & 47.12 & 65.01 \\
\textcolor{ForestGreen}{$\quad\Delta$} & \textcolor{ForestGreen}{-2.10} & \textcolor{ForestGreen}{-2.30} & \textcolor{ForestGreen}{-2.36} & \textcolor{ForestGreen}{-2.11} & \textcolor{ForestGreen}{-2.37} & \textcolor{ForestGreen}{-2.43} & \textcolor{ForestGreen}{-2.00} & \textcolor{ForestGreen}{-1.56} & \textcolor{ForestGreen}{-2.00} & \textcolor{ForestGreen}{-2.67} & \textcolor{ForestGreen}{-3.03} & \textcolor{ForestGreen}{-1.99} & \textcolor{ForestGreen}{-3.09} & \textcolor{ForestGreen}{-3.64} & \textcolor{ForestGreen}{-2.31} & \textcolor{ForestGreen}{-2.18} \\

\bottomrule
\end{tabular}}

\end{table*}

\section{Supplement Analysis}
In this section, we present supplementary analyses to further assess the robustness of our approach. We first conduct an ablation study to evaluate the contribution of each component in Section~\ref{sec:ablation}, followed by a convergence analysis to examine optimization stability in Section~\ref{sec:converagence}. We then analyze the effect of different initial prompts $P_0$ in Section~\ref{sec:diff_pro}, and the influence of sample size on performance in Section~\ref{sec:sample}.
\subsection{Ablation Study}
\label{sec:ablation} 

To better understand the internal mechanisms of the APOLO framework and evaluate the contribution of its core components, we conduct a systematic ablation study. We remove three essential modules, namely the Planner, the Socratic module, and the Critic, and examine their effects on performance across three representative LLM backbones. The results present in Table~\ref{tab:ablation} show that each module plays a vital role in maintaining APOLO’s effectiveness, with a clear hierarchy of importance among them.

\textbf{The Socratic module, which consists of the \textit{Teacher}, \textit{Student}, and \textit{Critic} agents, proves to be the most crucial component in emotion diagnosis task.} Removing it, referred to as ${w/o}_\text{Soc}$, leads to the most significant performance degradation. On average, Macro F1 decreases by 7.19\%, Micro F1 by 8.69\%, EMR by 11.22\%, and PMA by 12.07\%. This module acts as the optimization core of APOLO, driving iterative reflection and improvement through multi-round question–answer interactions. Without this process, APOLO loses its ability to iteratively refine prompts, reverting to a simple single-pass generator guided only by initial planning. The sharp decline in all metrics demonstrates that the dialog-based iterative refinement is not an auxiliary enhancement but the fundamental mechanism enabling APOLO to achieve its superior performance.

\textbf{The removal of the \textit{Planner} leads to the second most significant performance decline.} Specifically, ${w/o}_\text{Plan}$ results in average decreases of 3.75\% in Macro F1, 4.90\% in Micro F1, 6.66\% in EMR, and 6.73\% in PMA. The Planner decomposes complex optimization tasks into manageable subgoals, allowing the Socratic module to focus on improving specific aspects of the prompt. In its absence, the model must optimize the entire prompt at once, which is less efficient, less focused, and more prone to suboptimal reasoning paths. These results emphasize that the Planner plays a critical role in structuring the reasoning process and enabling efficient multi-stage refinement.

\textbf{The \textit{Critic} contributes the least to performance decline among all components.} Removing this agent (${w/o}_\text{Cri}$) leads to smaller but consistent performance declines, including 2.00\% in Macro F1, 3.00\% in Micro F1, 4.08\% in EMR, and 3.72\% in PMA. The Critic monitors the interaction between the Teacher and Student, ensuring that questions remain focused, logical, and open-ended. Without it, the dialogue may deviate from the intended principles, reducing the precision and consistency of the refinement process. These findings indicate that the Critic operates as a meta-optimization layer that ensures stable and principled reasoning throughout the iterative process.

\begin{figure*}[t]
	\large
	\centering 
	\includegraphics[scale=0.47]{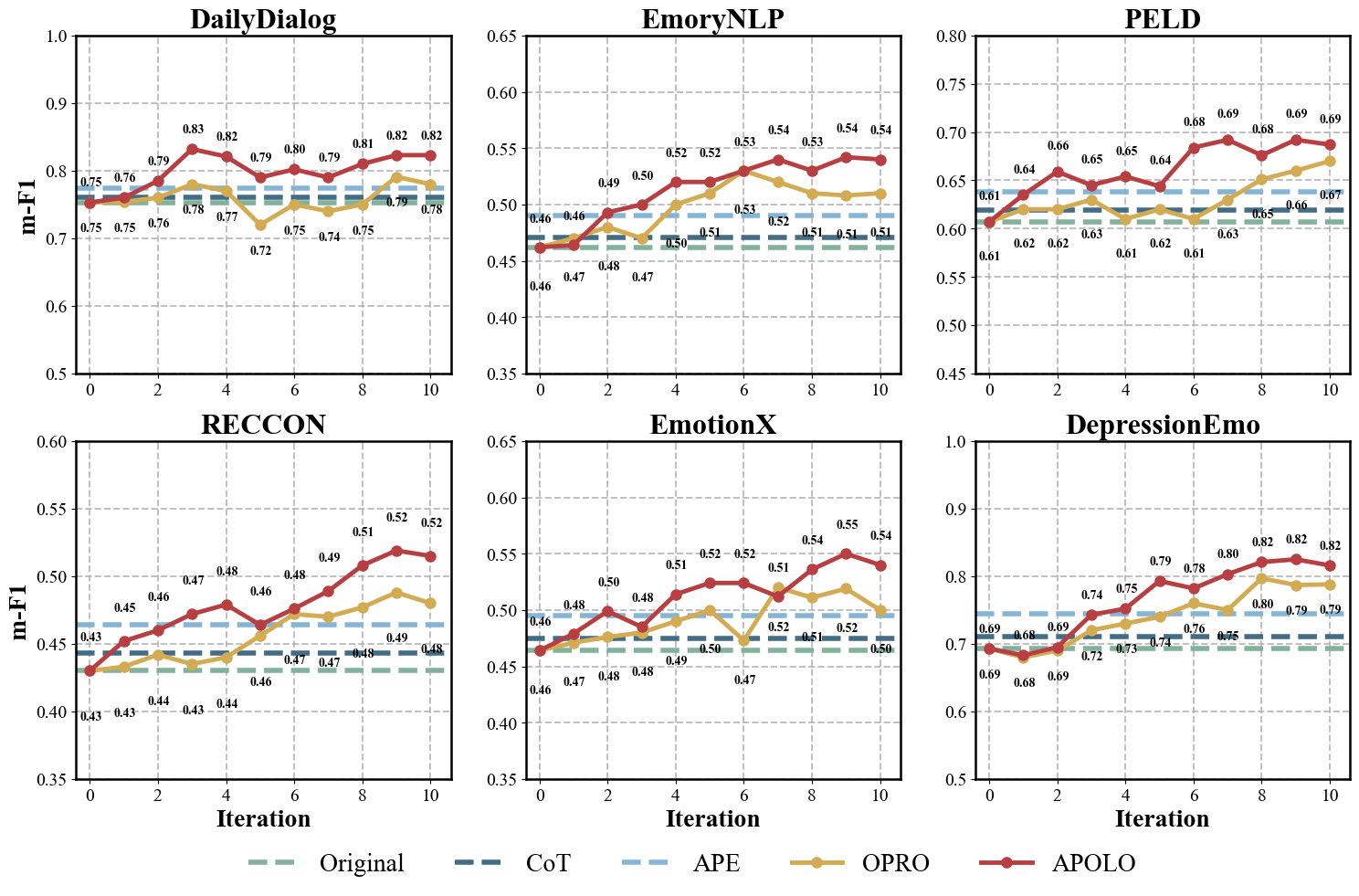}
	\caption{Convergence analysis of APOLO's optimization process on six emotion diagnosis datasets. The plots track the Micro F1-score (m-F1) over 10 iterations. APOLO's trajectory is compared against the iterative baseline OPRO and three static baselines (Original, CoT, APE).}
	\label{convergence}
\end{figure*}

\subsection{Converagence Analysis} 
\label{sec:converagence}
\textbf{APOLO demonstrates rapid and stable convergence in emotion diagnosis tasks, effectively balancing optimization efficiency and computational cost.}
Beyond achieving state-of-the-art performance, an essential property of an automated optimization framework for emotional reasoning lies in its efficiency. A practical method should not only discover high-quality diagnostic prompts but also do so with minimal time and computational overhead. To this end, we analyze APOLO’s optimization trajectory to examine its convergence behavior and efficiency characteristics.

As shown in Figure~\ref{convergence}, which plots validation performance (m-F1) against optimization iterations across six emotion-related datasets (DailyDialog, EmoryNLP, PELD, RECCON, EmotionX, and DepressionEmo), APOLO consistently exhibits a clear pattern of rapid early improvement followed by smooth stabilization. During the initial iterations (typically 1–4), APOLO achieves a sharp increase in m-F1, substantially outperforming all baselines as it rapidly corrects major emotional reasoning and alignment deficiencies in the initial prompt. This phase accounts for the majority of the total improvement, with early-stage gains ranging from approximately 5\% to 10\%.

After this surge, the performance gradually plateaus between iterations 6–10, reflecting a transition from coarse structural optimization to fine-grained affective refinement. In this stage, improvements become smaller yet more stable, indicating that the optimization focuses on subtle behavioral adjustments rather than large-scale modifications. Compared to OPRO, which improves more slowly and exhibits mild fluctuations, APOLO maintains a smoother and more monotonic convergence curve, particularly on RECCON and EmotionX where emotional convergence stability is typically harder to achieve.

\begin{table*}[t]
\caption{Final performance (Micro F1-scores) of APOLO on the \textit{RECCON} dataset (using \texttt{GPT-5-mini}) when initialized with six different starting prompts ($P_0$). The table shows the initial performance of each prompt and the final performance after optimization by APOLO.}
\label{tab:initial_prompts}
\centering
\small
\renewcommand{\arraystretch}{1.3} 
\begin{tabular}{@{}l|
>{\centering\arraybackslash}p{2.2cm} | >{\centering\arraybackslash}p{2.2cm} | >{\centering\arraybackslash}p{2.2cm} |
>{\centering\arraybackslash}p{2.2cm} | >{\centering\arraybackslash}p{2.2cm} | >{\centering\arraybackslash}p{2.2cm}@{}}
\toprule
\textbf{Method} & \textbf{Prompt 1} & \textbf{Prompt 2} & \textbf{Prompt 3} & \textbf{Prompt 4} & \textbf{Prompt 5} & \textbf{Prompt 6} \\
\midrule
\textbf{$P_0$} & \textbf{\textit{Choose the single most appropriate label for the utterance from the options.}} & \textit{What is the speaker's emotion? Choose from the options.} & \textit{You are an AI assistant. Your task is to identify the emotion.} & \textit{Your task is to output a single emotion for the utterance.} & \textit{Based on the conversation, identify the emotion.} & \textit{Please tell me the emotion of the last sentence.} \\
\midrule
Origin & 43.01 & 42.22 & 42.03 & 42.88 & 42.54 & 41.05 \\
\midrule
\textbf{APOLO} & \textbf{51.89} & \textbf{50.95} & \textbf{51.09} & \textbf{50.11} & \textbf{50.99} & \textbf{49.88} \\
\bottomrule
\end{tabular}

\end{table*}

\subsection{Different Initial Prompts $P_0$}
\label{sec:diff_pro}
\textbf{APOLO consistently converges to high-performing solutions in emotion diagnosis tasks regardless of the quality of the initial prompt, demonstrating the robustness of its optimization trajectory.} The starting point ($P_0$) can influence the path of optimization, so a truly robust framework should be able to find an effective solution even if the initial prompt is suboptimal. We test APOLO's resilience by initializing it with six distinct prompts of varying styles and quality on the \textit{RECCON} dataset. The prompts range from a simple, direct command to a question-based format and a basic role-playing instruction.

As shown in Table~\ref{tab:initial_prompts}, although the different starting prompts would lead to different initial performance levels, APOLO consistently guides the optimization towards a state of high performance. The final Micro F1-scores are all within a very narrow range, with a standard deviation of only \textbf{0.6647}, indicating that the framework is not highly sensitive to the quality of the initialization. This highlights APOLO's reliability: it can effectively adapt and refine its strategy to discover a superior prompt, even from a less-than-ideal starting point. This makes it a highly dependable tool for emotion-related applications for users, regardless of their initial prompt engineering expertise.

\subsection{Sample Size Analysis}
\label{sec:sample}
A critical challenge in developing robust models for linguistic emotion diagnosis is the inherent cost and scarcity of high-quality, expert-annotated data, particularly for nuanced tasks like those in the \textit{DepressionEmo} dataset. A practical framework must therefore be highly data-efficient. To rigorously evaluate APOLO's efficiency, we analyze its performance under zero-shot, one-shot, and three-shot settings, benchmarking it against baseline methods that rely on substantially larger sample sizes for their optimization processes.

The results, presented in Table~\ref{samples}, are unequivocal: APOLO, when provided with just a single exemplary dialogue (1-shot), consistently outperforms all baseline methods, even those that leverage up to 100 samples to guide their search. This is particularly evident on the complex \textit{DepressionEmo} dataset, where APOLO (1-shot) achieves a Micro F1-score of \textbf{82.45\%}, surpassing the strongest baseline, OPRO (which uses 50 samples). This demonstrates that APOLO can achieve state-of-the-art performance with a fraction of the data required by other leading frameworks.

Furthermore, the results show that increasing the sample size from one to three shots yields only marginal improvements across all six emotion diagnosis tasks. This indicates that a single, well-chosen example is sufficient for APOLO to capture the essential principles for effective prompting, making the one-shot configuration the best trade-off between performance and efficiency. This efficiency highlights the strength of APOLO’s Socratic refinement mechanism, which extracts deep and generalizable insights from minimal data. Consequently, APOLO emerges as both a high-performing and cost-effective framework for real-world emotion diagnosis, where labeled data is often scarce.

\begin{table}[t]
  \caption{Performance (\textbf{Micro F1-scores}) comparison of different sampling strategies on the evaluation metric across \textit{DailyDialog (D.D.), EmoryNLP (E.N.), PELD (PE.), RECCON (RE.), EmotionX (E.X), and DepressionEmo (D.E.)}, respectively. The experiments are conducted using the \texttt{GPT-5-mini} model. \textit{Train} means the training data.}
  \label{samples}
\setlength{\tabcolsep}{4pt}
  \centering
  \small 
  \begin{tabular}{cc|cccccc}
    \toprule
      \textbf{Tasks} & \textit{Train} & \textbf{D.D.} & \textbf{E.N.} & \textbf{PE.} & \textbf{RE.} & \textbf{E.X} & \textbf{D.E.} \\
    \midrule
      APE & 100 & 77.42 & 48.95 & 63.78 & 46.43 & 49.54 & 74.50 \\
      ProTeGi & 20 & 77.98 & 49.70 & 64.55 & 47.21 & 50.33 & 75.94 \\
      OPRO & 50 & 79.12 & 50.83 & 66.00 & 48.75 & 51.89 & 78.70 \\
      PE2 & 100 & 78.68 & 51.20 & 65.45 & 48.12 & 51.27 & 77.48 \\
    \midrule
      APOLO & 0 & 80.76 & 52.28 & 67.09 & 50.83 & 54.35 & 80.94 \\ 
      \textbf{APOLO} & \textbf{1} & \textbf{82.34} & \textbf{54.22} & \textbf{69.23} & \textbf{51.89} & \textbf{55.02} & \textbf{82.45} \\ 
      APOLO & 3 & 82.57 & 54.69 & 69.30 & 51.24 & 55.96 & 83.35 \\ 
    \bottomrule
  \end{tabular} 

\end{table}

\section{Details of Datasets and  Optimized Prompts \label{sec:details}}
In this section, we first introduce the public datasets used in our experiments on linguistic emotion diagnosis in Section~\ref{sec:dataset}. We then describe the system prompts that define the roles and behaviors of each agent in Section~\ref{sec:prompt}. Finally, we present a case study illustrating one complete optimization iteration in Section~\ref{sec:case}.

\subsection{Datasets}
\label{sec:dataset}

\textbf{DepressionEmo~\cite{rahman2024depressionemo}.}
The DepressionEmo dataset is a multi-label corpus that captures a broad range of depressive emotions expressed through naturalistic online text. It comprises user-generated posts from online communities, annotated with multiple emotional dimensions such as sadness, hopelessness, loneliness, worthlessness, and emptiness. Unlike conventional emotion classification datasets that focus on basic affective states, DepressionEmo provides fine-grained distinctions among depressive manifestations, making it suitable for tasks that involve psychological or clinical language patterns. In this work, it serves as an essential benchmark to evaluate whether the proposed prompt optimization framework can effectively diagnose and disentangle overlapping emotional cues in long, unstructured text typical of mental health discourse.

\textbf{DailyDialog~\cite{li2017dailydialog}.}
The DailyDialog dataset contains multi-turn dialogues representing various everyday conversational situations, each labeled with both emotion and communicative intent. The test split used in this study includes 1,000 dialogues, encompassing a diverse set of topics and interpersonal scenarios. This dataset provides high-quality human-written language with well-formed sentences and clear emotion shifts across turns, such as transitions between happiness, anger, sadness, and surprise. We employ it to examine how well our optimized prompts generalize to open-domain, emotionally varied conversations. Its structured dialogues also allow for consistent evaluation of linguistic sensitivity, ensuring that improvements in emotion diagnosis stem from the prompt design rather than noise or lexical imbalance.

\textbf{EmoryNLP~\cite{zahiri2018emorynlp}.}
For the EmoryNLP collection we specifically employ the Emotion Detection subdataset, released in May 2017. This subdataset provides categorical emotion labels for utterances drawn from multiparty scripted dialogues, preserving speaker identity and conversational context that are essential for analyzing emotion flow in multi-participant interactions. In our experiments we use the official test split of this subdataset, which contains 1,328 utterances annotated across seven categorical emotion labels. Because the material is dialogic and speaker-anchored, it is particularly useful for assessing whether our prompt optimization method can capture speaker-specific affective patterns and fine-grained contextual emotion cues in multiparty exchanges.

\textbf{PELD (Personality EmotionLines Dataset)~\cite{peld}.}
The PELD dataset extends the standard emotion recognition setting by integrating personality factors with emotional annotations in dialogue. It comprises over six thousand dialogue triples, each annotated with both the speaker’s personality profile, based on the Big Five dimensions, and the expressed emotion of the response. By combining personality and emotion, PELD captures how individual differences shape linguistic manifestations of affect. This dataset allows us to evaluate the adaptability of our prompt optimization framework to personality-driven variation in emotion expression, providing a more realistic and psychologically grounded testing environment for linguistic emotion diagnosis.

\begin{table*}[t]
\caption{Prompts for all Agents in APOLO framework. The examples in the table are from the Emotion Diagnosis Task of the \textit{DepressionEmo} dataset.}
\label{agentPrompts}
  \renewcommand{\arraystretch}{1.2} 
  \centering
  \small
  \begin{tabular}{p{18cm}} 
  \toprule
  \textbf{\textit{Planner}} \\
    You are a medical emotion diagnosis planning assistant.
    Your role is to generate an efficient and structured plan to solve the next problem related to medical emotion diagnosis (for example, identifying emotional states from patient text, integrating physiological data, or improving diagnostic accuracy).
    Your plan should be goal-oriented, concise, and executable, reflecting logical reasoning and domain awareness.

Your response must strictly follow the format below:
Total steps: [number]
Step 1: [description]
Step 2: [description] 
...
Step N: [description]

For example:
Total steps: 2
Step 1: Review the emotional indicators described in the problem.
Step 2: Select the appropriate diagnostic model or analytical method.

Please follow this format strictly.
\\

  \midrule
  \textbf{\textit{Teacher}} \\
    You are a teacher who asks questions in the Socratic manner, focusing on medical emotion diagnosis tasks.
    Your goal is to help students think critically about identifying, interpreting, or improving emotional diagnosis in medical contexts.
    Please ask a total of two questions:
    The first one is for the problem that appeared in the prompt given by the students in the last round.
    The second one is an optimization solution based on the current steps of the task.

Please include only questions in your output and do not make answers for your students.\\

  \midrule
  \textbf{\textit{Student}} \\
    You are a prompt generator specialized in medical emotion diagnosis.
    Your task is to iterate over existing prompts and refine them so they better address tasks related to identifying, interpreting, and reasoning about emotional states in medical contexts (such as analyzing patient narratives, clinical notes, or multimodal emotional indicators).
    Focus on improving the prompt’s clarity, diagnostic relevance, and reasoning efficiency, ensuring that the resulting prompt effectively guides agents to generate insights or actions beneficial for medical emotion diagnosis.
    
    Note: Output only the newly generated prompt and nothing else.\\
  \midrule
  \textbf{\textit{Critic}} \\
  You are an evaluator specialized in medical emotion diagnosis, responsible for judging the correctness and clinical relevance of a given task or response related to identifying, analyzing, or reasoning about emotional states in medical contexts.
  Your output must strictly follow these rules:

1. If the task is judged as correct, output only:
   [True]

2. If the task is judged as incorrect, output:
   [False]
   [suggestion: reason for the incorrect judgment]

   Replace `reason for the incorrect judgment` with a clear and concise explanation of why the task is incorrect.

Do not include any additional text, comments, or explanations beyond the specified format.\\
  \midrule
  \textbf{\textit{Target}} \\
    Analyze the given user's post carefully and identify all relevant psychological emotion labels that accurately reflect the mental health state expressed in the text. Follow these steps:  
    
    1. **Read the post thoroughly**: Pay close attention to both the title and main content to understand the full context.  
    2. **Identify key emotional cues**: Look for words, phrases, or tones that indicate specific emotions (e.g., 'I feel like a waste of space' suggests worthlessness).  
    3. **Match cues to emotion labels**: Compare the identified cues with the provided emotion labels to determine which ones fit best.  
    4. **Consider overlapping emotions**: Some posts may express multiple emotions—ensure all applicable labels are selected.  
    5. **Justify your selections**: Briefly explain why each chosen label is appropriate based on the text.  
    
    Example Output Format:  
    **Emotions**: [worthlessness, hopelessness, sadness]  
    **Reasoning**: The post expresses feelings of inadequacy ('waste of space'), lack of motivation ('hard to find motivation'), and persistent sadness ('dislike everything about yourself').  
    
    Now, analyze the following post and provide the most accurate emotion labels:  
    [Insert Post Here]  
    Options: [anger, brain dysfunction (forget), emptiness, hopelessness, loneliness, sadness, suicide intent, worthlessness]
    --- Conversation History ---
     --- Current Utterance to Analyze Your output MUST be a single word representing the emotion. Do not provide any explanation or other text."
\\
  \bottomrule
  \end{tabular}

\end{table*}

\textbf{RECCON~\cite{poria2021recognizing}.}
The RECCON dataset, which aims at Recognizing Emotion Cause in CONversations, frames conversational emotion analysis as a causality-aware task: beyond labeling the emotion of a target utterance, it requires identification or validation of the antecedent contextual elements that plausibly triggered that emotion. This dual focus, emotion recognition plus cause identification, elevates the task from surface classification to context-sensitive interpretation, demanding reasoning over discourse structure and pragmatic cues. We include RECCON to probe whether prompt optimization can improve not only categorical emotion recognition but also the model’s ability to associate emotional states with their linguistic antecedents, thereby supporting more interpretable and discourse-aware linguistic emotion diagnosis.

\textbf{EmotionX (EmotionX-2019)~\cite{shmueli2019socialnlpemotionx2019challenge}.}
For EmotionX we utilize the 2018\_EmotionX subdataset, specifically the test partitions drawn from two complementary sources: the Friends transcripts and the EmotionPush chat logs. In our setup we evaluate on the Friends test portion comprising 2,764 sentences and the EmotionPush test portion comprising 2,807 sentences. Each utterance in these partitions is labeled with one of seven basic emotion categories (neutral, joy, sadness, fear, anger, surprise, disgust). Because the two subcorpora differ in register, scripted televisual dialogue versus spontaneous chat, the combined evaluation enables an analysis of how prompt optimization affects robustness across distinct dialogue domains, stylistic registers, and emotion expression modalities.

\begin{figure*}[!t]
    \centering
    \includegraphics[scale=0.32]{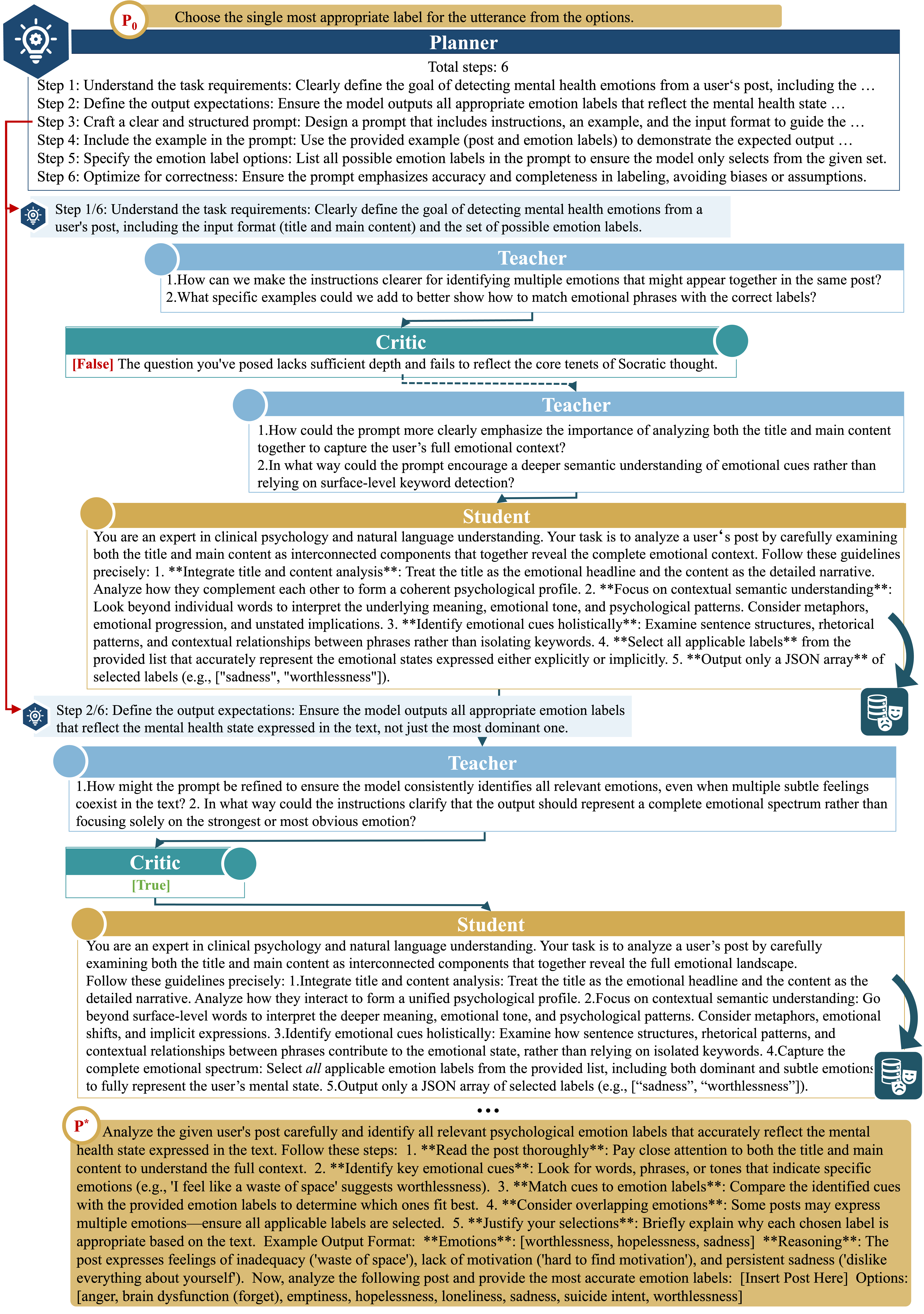}
    \caption{A complete example of the collaborative output from all agents in a single iteration, using the Emotion Diagnosis Task of the \textit{DepressionEmo} dataset.}
    \label{fullSteps}
\end{figure*}

\subsection{Prompts for Agents}
\label{sec:prompt}

The behavior of the APOLO framework is primarily governed by the system prompts assigned to its agents. These prompts define each agent’s role, constraints, and communication protocols, ensuring a structured and effective optimization process. Table~\ref{agentPrompts} summarizes the exact meta-prompts used for the \textit{Planner} and the Socratic module (\textit{Teacher}, \textit{Student}, and \textit{Critic}), along with an example of a final optimized prompt produced for the \textit{Target} agent.

The \textit{Planner} prompt emphasizes structure and reliability, enforcing a strict template that decomposes the optimization task into clear and actionable steps, thereby ensuring consistent and machine-parsable outputs that effectively guide the subsequent Socratic refinement phase. The Socratic module, which consists of the \textit{Teacher}, \textit{Student}, and \textit{Critic}, is designed to facilitate a goal-oriented yet exploratory dialogue. Within this module, the \textit{Teacher} poses probing and constructive 
questions, the \textit{Student} iteratively refines its responses based on the feedback, and the \textit{Critic} evaluates the exchange by providing binary judgments with concise justifications to preserve rigor and focus. Together, these components form the core mechanism that drives APOLO’s iterative optimization process.


In contrast, the \textit{Target} agent’s prompt represents the final output of the entire APOLO process rather than a predefined instruction. The example shown for the DepressionEmo dataset illustrates the end result of APOLO’s refinement: a structured, multi-step instruction that integrates explicit role guidance (“Analyze the given user’s post carefully”), a clear Chain-of-Thought reasoning sequence (Steps 1–5), and precise output formatting. This transformation from a simple initial command to a sophisticated, well-structured instruction captures the essence of APOLO’s methodology and explains its superior empirical performance.

\subsection{Case Study}
\label{sec:case}
To provide a concrete illustration of the APOLO framework's inner workings, we present a step-by-step case study of one complete optimization iteration. Figure~\ref{fullSteps} visualizes this entire collaborative process, from the initial prompt ($P_0$) to the refined prompt generated at the end of the Socratic dialogue, using the challenging multi-label Emotion Diagnosis Task from the DepressionEmo dataset.

The process begins with a simple and somewhat ambiguous initial prompt, $P_0$: ``Choose the single most appropriate label for the utterance from the options.'' Recognizing the complexity of the task, the \textbf{\textit{Planner}} agent first decomposes the optimization goal into a structured, six-step plan. This plan includes crucial sub-goals such as defining the output expectations (Step 2) and ensuring the prompt handles multiple emotions (Step 2), which immediately addresses the shortcomings of the initial prompt.

The framework then proceeds to execute this plan step-by-step. For \textbf{Step 1}, "Understand the task requirements," the \textbf{\textit{Teacher}} poses a question to the \textbf{\textit{Student}} aimed at improving the clarity of the instructions. However, the \textbf{\textit{Critic}} intervenes, judging the Teacher's initial question as ``[False]'' because it lacks sufficient depth and fails to adhere to Socratic principles. This feedback loop is critical; it forces the Teacher to re-evaluate and ask a more profound, open-ended question: ``How could the prompt more clearly emphasize the importance of analyzing both the title and main content...?''

In response to this improved Socratic inquiry, the \textbf{\textit{Student}} generates a significantly more sophisticated prompt. It now includes explicit role-playing ("You are an expert in clinical psychology...") and detailed guidelines, such as integrating title and content analysis and focusing on contextual semantic understanding. This process repeats for subsequent steps. For \textbf{Step 2}, "Define the output expectations," the Teacher's question (approved by the Critic with ``[True]'') prompts the Student to further refine the prompt to handle multiple coexisting emotions.

This iterative, multi-agent dialogue continues until all of the Planner's steps are completed. The final output, denoted as $P^*$, is a vastly improved prompt that incorporates a Chain-of-Thought structure, role-playing, and precise output constraints. This walkthrough clearly demonstrates how the structured planning, Socratic refinement, and critical feedback mechanisms in APOLO collaborate to systematically transform a naive initial prompt into a highly effective, nuanced instruction.


\section{Conclusion}

In this work, we address two fundamental challenges in disease-related emotion recognition: 
\textbf{emotional comorbidity} and \textbf{inefficient exploration}. 
To tackle these issues, we propose \textbf{APOLO}, 
a POMDP-based multi-agent framework that enables dynamic, interpretable, and risk-aware prompt optimization. 
By incorporating the collaborative roles of \textit{Planner–Teacher–Critic–Student–Target}, 
APOLO systematically explores the prompt space through adaptive reasoning and feedback alignment, 
achieving more comprehensive emotion coverage with enhanced stability and interpretability. 
The framework mitigates emotional comorbidity via adaptive semantic reasoning and improves exploration efficiency 
through uncertainty modeling and multi-agent collaboration. 
Extensive experiments on multilingual and risk-sensitive benchmarks demonstrate that APOLO 
consistently outperforms existing methods in diagnostic accuracy, calibration, and robustness. 
This work provides a scalable and trustworthy paradigm for leveraging large language models in mental health assessment and clinical decision support.





\nocite{*}

\bibliographystyle{IEEEtran}
\bibliography{Draft}

\begin{thebibliography}{10}
\providecommand{\url}[1]{#1}
\csname url@samestyle\endcsname
\providecommand{\newblock}{\relax}
\providecommand{\bibinfo}[2]{#2}
\providecommand{\BIBentrySTDinterwordspacing}{\spaceskip=0pt\relax}
\providecommand{\BIBentryALTinterwordstretchfactor}{4}
\providecommand{\BIBentryALTinterwordspacing}{\spaceskip=\fontdimen2\font plus
\BIBentryALTinterwordstretchfactor\fontdimen3\font minus \fontdimen4\font\relax}
\providecommand{\BIBforeignlanguage}[2]{{%
\expandafter\ifx\csname l@#1\endcsname\relax
\typeout{** WARNING: IEEEtran.bst: No hyphenation pattern has been}%
\typeout{** loaded for the language `#1'. Using the pattern for}%
\typeout{** the default language instead.}%
\else
\language=\csname l@#1\endcsname
\fi
#2}}
\providecommand{\BIBdecl}{\relax}
\BIBdecl

\bibitem{kumar2024computational}
P.~Kumar, A.~Vedernikov, Y.~Chen, W.~Zheng, and X.~Li, ``Computational analysis of stress, depression and engagement in mental health: A survey,'' \emph{arXiv preprint arXiv:2403.08824}, 2024.

\bibitem{ge2025survey}
Z.~Ge, N.~Hu, D.~Li, Y.~Wang, S.~Qi, Y.~Xu, H.~Shi, and J.~Zhang, ``A survey of large language models in mental health disorder detection on social media,'' in \emph{2025 IEEE 41st International Conference on Data Engineering Workshops (ICDEW)}.\hskip 1em plus 0.5em minus 0.4em\relax IEEE, 2025, pp. 164--176.

\bibitem{zhang-physreason}
\BIBentryALTinterwordspacing
X.~Zhang, Y.~Dong, Y.~Wu, J.~Huang, C.~Jia, B.~Fernando, M.~Z. Shou, L.~Zhang, and J.~Liu, ``{P}hys{R}eason: A comprehensive benchmark towards physics-based reasoning,'' in \emph{Proceedings of the 63rd Annual Meeting of the Association for Computational Linguistics (Volume 1: Long Papers)}.\hskip 1em plus 0.5em minus 0.4em\relax Association for Computational Linguistics, 2025, pp. 16\,593--16\,615. [Online]. Available: \url{https://aclanthology.org/2025.acl-long.811/}
\BIBentrySTDinterwordspacing

\bibitem{bucur2025survey}
A.-M. Bucur, M.~Zampieri, T.~Ranasinghe, and F.~Crestani, ``A survey on multilingual mental disorders detection from social media data,'' \emph{arXiv preprint arXiv:2505.15556}, 2025.

\bibitem{zhang2023emotion}
T.~Zhang, K.~Yang, S.~Ji, and S.~Ananiadou, ``Emotion fusion for mental illness detection from social media: A survey,'' \emph{Information Fusion}, vol.~92, pp. 231--246, 2023.

\bibitem{fu2026erreval}
W.~Fu, B.~Wei, J.~Hao, Y.~Zhang, J.~Zhang, J.~Wang, B.~Li, Y.~He, L.~Zhang, and J.~Liu, ``Erreval: Error-aware evaluation for question generation through explicit diagnostics,'' \emph{arXiv preprint arXiv:2601.10406}, 2026.

\bibitem{yan2025mur}
H.~Yan, F.~Xu, R.~Xu, Y.~Li, J.~Zhang, H.~Luo, X.~Wu, L.~A. Tuan, H.~Zhao, Q.~Lin \emph{et~al.}, ``Mur: Momentum uncertainty guided reasoning for large language models,'' \emph{arXiv preprint arXiv:2507.14958}, 2025.

\bibitem{sahoo2024systematic}
P.~Sahoo, A.~K. Singh, S.~Saha, V.~Jain, S.~Mondal, and A.~Chadha, ``A systematic survey of prompt engineering in large language models: Techniques and applications,'' \emph{arXiv preprint arXiv:2402.07927}, 2024.

\bibitem{xu2025large}
F.~Xu, Q.~Lin, J.~Han, T.~Zhao, J.~Liu, and E.~Cambria, ``Are large language models really good logical reasoners? a comprehensive evaluation and beyond,'' \emph{IEEE Transactions on Knowledge and Data Engineering}, 2025.

\bibitem{lin2025has}
Q.~Lin, Y.~Zhu, X.~Mei, L.~Huang, J.~Ma, K.~He, Z.~Peng, E.~Cambria, and M.~Feng, ``Has multimodal learning delivered universal intelligence in healthcare? a comprehensive survey,'' \emph{Information Fusion}, vol. 116, p. 102795, 2025.

\bibitem{li2025survey}
W.~Li, X.~Wang, W.~Li, and B.~Jin, ``A survey of automatic prompt engineering: An optimization perspective,'' \emph{arXiv preprint arXiv:2502.11560}, 2025.

\bibitem{cui2025automatic}
W.~Cui, J.~Zhang, Z.~Li, H.~Sun, D.~Lopez, K.~Das, B.~A. Malin, and S.~Kumar, ``Automatic prompt optimization via heuristic search: A survey,'' \emph{arXiv preprint arXiv:2502.18746}, 2025.

\bibitem{ramnath2025systematic}
K.~Ramnath, K.~Zhou, S.~Guan, S.~S. Mishra, X.~Qi, Z.~Shen, S.~Wang, S.~Woo, S.~Jeoung, Y.~Wang \emph{et~al.}, ``A systematic survey of automatic prompt optimization techniques,'' \emph{arXiv preprint arXiv:2502.16923}, 2025.

\bibitem{chang2024efficient}
K.~Chang, S.~Xu, C.~Wang, Y.~Luo, X.~Liu, T.~Xiao, and J.~Zhu, ``Efficient prompting methods for large language models: A survey,'' \emph{arXiv preprint arXiv:2404.01077}, 2024.

\bibitem{hegde2025emotions}
K.~Hegde and H.~Jayalath, ``Emotions in the loop: A survey of affective computing for emotional support,'' \emph{arXiv preprint arXiv:2505.01542}, 2025.

\bibitem{zhou2023large}
Y.~Zhou, A.~I. Muresanu, Z.~Han, K.~Paster, S.~Pitis, H.~Chan, and J.~Ba, ``Large language models are human-level prompt engineers,'' in \emph{The Eleventh International Conference on Learning Representations (ICLR)}, 2023.

\bibitem{xureprompting}
W.~Xu, A.~Banburski-Fahey, and N.~Jojic, ``Reprompting: Automated chain-of-thought prompt inference through gibbs sampling,'' \emph{arXiv preprint arXiv:2305.09993}, 2023.

\bibitem{wang2023promptagent}
X.~Wang, C.~Li, Z.~Wang, F.~Bai, H.~Luo, J.~Zhang, N.~Jojic, E.~P. Xing, and Z.~Hu, ``Promptagent: Strategic planning with language models enables expert-level prompt optimization,'' \emph{arXiv preprint arXiv:2310.16427}, 2023.

\bibitem{Yang0LLLZC24}
\BIBentryALTinterwordspacing
C.~Yang, X.~Wang, Y.~Lu, H.~Liu, Q.~V. Le, D.~Zhou, and X.~Chen, ``Large language models as optimizers,'' in \emph{The Twelfth International Conference on Learning Representations, {ICLR} 2024, Vienna, Austria, May 7-11, 2024}.\hskip 1em plus 0.5em minus 0.4em\relax OpenReview.net, 2024. [Online]. Available: \url{https://openreview.net/forum?id=Bb4VGOWELI}
\BIBentrySTDinterwordspacing

\bibitem{ye2023prompt}
Q.~Ye, M.~Axmed, R.~Pryzant, and F.~Khani, ``Prompt engineering a prompt engineer,'' \emph{arXiv preprint arXiv:2311.05661}, 2023.

\bibitem{zhang2026mars}
J.~Zhang, Z.~Wang, H.~Zhu, J.~Liu, Q.~Lin, and E.~Cambria, ``Mars: A multi-agent framework incorporating socratic guidance for automated prompt optimization,'' in \emph{Proceedings of the AAAI Conference on Artificial Intelligence}, 2026.

\bibitem{singh2023detailed}
N.~Singh and U.~C. Jaiswal, ``A detailed sentiment analysis survey based on machine learning techniques,'' \emph{ADC-AIJ: Advances in Distributed Computing and Artificial Intelligence Journal}, vol.~12, no.~2, pp. 201--216, 2023.

\bibitem{hassan2023comprehensive}
A.~Hassan and M.~R. Islam, ``A comprehensive survey on sentiment analysis techniques,'' \emph{International Journal of Computers and Applications}, vol. 185, no.~12, pp. 1--11, 2023.

\bibitem{he2022jcbie}
K.~He, R.~Mao, T.~Gong, E.~Cambria, and C.~Li, ``Jcbie: a joint continual learning neural network for biomedical information extraction,'' \emph{BMC bioinformatics}, vol.~23, no.~1, p. 549, 2022.

\bibitem{zhu2024enhancing}
X.~Zhu, T.~Liu, Y.~Wu, and Z.~Wang, ``Enhancing facial emotion recognition through deep learning: Integrating {CNN} and {RNN-LSTM} models,'' in \emph{Proceedings of the 2nd International Conference on Machine Learning and Gerontechnology}, 2024, pp. 1--6.

\bibitem{bhat2021cnn}
V.~Bhat, R.~Shah, and N.~Mehendale, ``A {CNN-LSTM} based deep neural networks for facial emotion detection in videos,'' \emph{International Journal of Creative Research Thoughts (IJCRT)}, vol.~9, no.~11, pp. d327--d332, 2021.

\bibitem{mishra2020speech}
P.~Mishra and M.~P, ``Speech emotion recognition using {LSTM} and {RNN},'' \emph{Journal of Electrical Engineering}, vol.~20, no.~3, pp. 1--6, 2020.

\bibitem{lan2025gem}
X.~Lan, F.~Wu, K.~He, Q.~Zhao, S.~Hong, and M.~Feng, ``Gem: Empowering mllm for grounded ecg understanding with time series and images,'' \emph{arXiv preprint arXiv:2503.06073}, 2025.

\bibitem{acheampong2021transformer}
F.~A. Acheampong, H.~Nunoo-Mensah, and W.~Chen, ``Transformer models for text-based emotion detection: a review of {BERT}-based approaches,'' \emph{Artificial Intelligence Review}, vol.~54, no.~8, pp. 5789--5829, 2021.

\bibitem{bao2021bert}
H.~Bao, K.~He, X.~Yin, X.~Li, X.~Bao, H.~Zhang, J.~Wu, and Z.~Gao, ``Bert-based meta-learning approach with looking back for sentiment analysis of literary book reviews,'' in \emph{CCF International Conference on Natural Language Processing and Chinese Computing}.\hskip 1em plus 0.5em minus 0.4em\relax Springer, 2021, pp. 235--247.

\bibitem{vishnubhotla2023language}
K.~Vishnubhotla and S.~M. Mohammad, ``Language and mental health: Measures of emotion dynamics from text as linguistic biosocial markers,'' in \emph{Proceedings of the 2023 Conference on Empirical Methods in Natural Language Processing (EMNLP)}.\hskip 1em plus 0.5em minus 0.4em\relax Association for Computational Linguistics, 2023, pp. 3117--3133.

\bibitem{shin2020autoprompt}
T.~Shin, Y.~Razeghi, R.~L. Logan~IV, E.~Wallace, and S.~Singh, ``{AutoPrompt}: Eliciting knowledge from language models with automatically generated prompts,'' in \emph{Proceedings of the 2020 Conference on Empirical Methods in Natural Language Processing (EMNLP)}.\hskip 1em plus 0.5em minus 0.4em\relax Association for Computational Linguistics, 2020, pp. 4222--4235.

\bibitem{li2021prefix}
X.~L. Li and P.~Liang, ``{Prefix-Tuning}: Optimizing continuous prompts for generation,'' in \emph{Proceedings of the 59th Annual Meeting of the Association for Computational Linguistics and the 11th International Joint Conference on Natural Language Processing (Volume 1: Long Papers)}.\hskip 1em plus 0.5em minus 0.4em\relax Association for Computational Linguistics, 2021, pp. 4582--4597.

\bibitem{zhang2025gkg}
J.~Zhang, S.~Qi, Y.~Dong, L.~Yuan, T.~Shen, W.~Fu, B.~Wei, H.~Zhu, and J.~Liu, ``Gkg-llm: A unified framework for generalized knowledge graph construction,'' \emph{Information Fusion}, p. 103956, 2025.

\bibitem{zhang2026maxs}
J.~Zhang, Z.~Wang, Z.~Wang, Y.~He, H.~Luo, L.~Zhang, R.~Mao, Q.~Lin, J.~Liu \emph{et~al.}, ``Maxs: Meta-adaptive exploration with llm agents,'' \emph{arXiv preprint arXiv:2601.09259}, 2026.

\bibitem{zhang2024meta}
\BIBentryALTinterwordspacing
Y.~Zhang, Y.~Yuan, and A.~C.-C. Yao, ``Meta prompting for ai systems,'' 2025. [Online]. Available: \url{https://arxiv.org/abs/2311.11482}
\BIBentrySTDinterwordspacing

\bibitem{deng2022rlprompt}
M.~Deng, J.~Wang, C.-P. Hsieh, Y.~Wang, H.~Guo, T.~Shu, M.~Song, E.~P. Xing, and Z.~Hu, ``{RLPrompt}: Optimizing discrete text prompts with reinforcement learning,'' in \emph{Proceedings of the 2022 Conference on Empirical Methods in Natural Language Processing (EMNLP)}.\hskip 1em plus 0.5em minus 0.4em\relax Association for Computational Linguistics, 2022, pp. 7576--7593.

\bibitem{zhang2025maps}
J.~Zhang, Z.~Wang, Z.~Wang, X.~Zhang, F.~Xu, Q.~Lin, R.~Mao, E.~Cambria, and J.~Liu, ``Maps: A multi-agent framework based on big seven personality and socratic guidance for multimodal scientific problem solving,'' \emph{arXiv preprint arXiv:2503.16905}, 2025.

\bibitem{li2017dailydialog}
Y.~Li, H.~Su, X.~Shen, W.~Li, Z.~Cao, and S.~Niu, ``Dailydialog: A manually labelled multi-turn dialogue dataset,'' \emph{arXiv preprint arXiv:1710.03957}, 2017.

\bibitem{zahiri2018emorynlp}
S.~M. Zahiri and J.~D. Choi, ``Emotion detection on tv show transcripts with sequence-based convolutional neural networks.'' in \emph{AAAI Workshops}, vol.~18, 2018, pp. 44--52.

\bibitem{peld}
Z.~Wen, J.~Cao, R.~Yang, S.~Liu, and J.~Shen, ``Automatically select emotion for response via personality-affected emotion transition,'' in \emph{Findings of the Association for Computational Linguistics: ACL-IJCNLP 2021}, 2021, pp. 5010--5020.

\bibitem{poria2021recognizing}
S.~Poria, N.~Majumder, D.~Hazarika, D.~Ghosal, R.~Bhardwaj, S.~Y.~B. Jian, P.~Hong, R.~Ghosh, A.~Roy, N.~Chhaya \emph{et~al.}, ``Recognizing emotion cause in conversations,'' \emph{Cognitive Computation}, vol.~13, no.~5, pp. 1317--1332, 2021.

\bibitem{shmueli2019socialnlpemotionx2019challenge}
\BIBentryALTinterwordspacing
B.~Shmueli and L.-W. Ku, ``Socialnlp emotionx 2019 challenge overview: Predicting emotions in spoken dialogues and chats,'' 2019. [Online]. Available: \url{https://arxiv.org/abs/1909.07734}
\BIBentrySTDinterwordspacing

\bibitem{rahman2024depressionemo}
A.~B.~S. Rahman, H.-T. Ta, L.~Najjar, A.~Azadmanesh, and A.~S. G{\"o}nul, ``Depressionemo: A novel dataset for multilabel classification of depression emotions,'' \emph{Journal of Affective Disorders}, vol. 366, pp. 445--458, 2024.

\bibitem{openai_gpt5}
\BIBentryALTinterwordspacing
OpenAI. (2025) Introducing gpt-5. Accessed: Aug. 2025. [Online]. Available: \url{https://openai.com/index/introducing-gpt-5/}
\BIBentrySTDinterwordspacing

\bibitem{liu2024deepseek}
A.~Liu, B.~Feng, B.~Xue, B.~Wang, B.~Wu, C.~Lu, C.~Zhao, C.~Deng, C.~Zhang, C.~Ruan \emph{et~al.}, ``Deepseek-v3 technical report,'' \emph{arXiv preprint arXiv:2412.19437}, 2024.

\bibitem{yang2025qwen3}
A.~Yang, A.~Li, B.~Yang, B.~Zhang, B.~Hui, B.~Zheng, B.~Yu, C.~Gao, C.~Huang, C.~Lv \emph{et~al.}, ``Qwen3 technical report,'' \emph{arXiv preprint arXiv:2505.09388}, 2025.

\bibitem{pryzant2023automatic}
\BIBentryALTinterwordspacing
R.~Pryzant, D.~Iter, J.~Li, Y.~T. Lee, C.~Zhu, and M.~Zeng, ``Automatic prompt optimization with "gradient descent" and beam search,'' in \emph{Proceedings of the 2023 Conference on Empirical Methods in Natural Language Processing, (EMNLP)}, 2023, pp. 7957--7968. [Online]. Available: \url{https://doi.org/10.18653/v1/2023.emnlp-main.494}
\BIBentrySTDinterwordspacing

\bibitem{yang2024harnessing}
J.~Yang, H.~Jin, R.~Tang, X.~Han, Q.~Feng, H.~Jiang, S.~Zhong, B.~Yin, and X.~Hu, ``Harnessing the power of llms in practice: A survey on chatgpt and beyond,'' \emph{ACM Transactions on Knowledge Discovery from Data}, vol.~18, no.~6, pp. 1--32, 2024.

\bibitem{al-taei2023survey}
M.~A. Al-Taei and S.~M. Al-Taei, ``A survey of sentiment analysis: Approaches, datasets, and future research,'' \emph{Applied Sciences}, vol.~13, no.~12, p. 7091, 2023.

\bibitem{li2025prompt}
Z.~Li, Y.~Liu, Y.~Su, and N.~Collier, ``Prompt compression for large language models: A survey,'' in \emph{Proceedings of the 2025 Conference of the Nations of the Americas Chapter of the Association for Computational Linguistics: Human Language Technologies (Volume 1: Long Papers)}, 2025, pp. 7182--7195.

\end{thebibliography}

\end{document}